\documentclass[dvipsnames, svgnames, x11names, table]{article}
\PassOptionsToPackage{}{natbib}  

\usepackage{conf/colm2024_conference}
\colmfinalcopy  

\usepackage{color}
\usepackage{xcolor}

\definecolor{darkblue}{rgb}{0, 0, 0.5}
\definecolor{darkgreen}{rgb}{0, 0.5, 0}

\usepackage{CJKutf8} 
\usepackage{microtype}
\usepackage{graphicx}
\usepackage{subcaption}
\usepackage{booktabs}
\usepackage[
    colorlinks,
    linkcolor=darkblue,
    citecolor=darkblue,
    linkcolor=darkblue,
    urlcolor=darkblue
]{hyperref}
\usepackage{amsmath}
\usepackage{amssymb}
\usepackage{mathtools}
\usepackage{amsthm}
\usepackage[capitalize,noabbrev]{cleveref}

\usepackage[utf8]{inputenc}
\usepackage[T1]{fontenc}
\usepackage{url}
\usepackage{amsfonts}
\usepackage{nicefrac}
\usepackage{latexsym}
\usepackage{array}
\usepackage{multirow}
\usepackage{alltt}
\usepackage{enumerate}
\usepackage{bbm}
\usepackage{eucal}
\usepackage{xspace}
\usepackage{stmaryrd}
\usepackage{bussproofs}
\usepackage{mathpartir}
\usepackage{tikz}
\usepackage{pgfmath}
\usepackage{xurl}
\usepackage{mdframed}
\usepackage{enumitem}
\usepackage{tabularx}
\usepackage{listings}
\usepackage{xparse}
\usepackage{xifthen}
\usepackage{multicol}
\usepackage{xfrac}
\usepackage{appendix}
\usepackage{etoolbox}
\usepackage{floatflt}
\usepackage{wrapfig}
\usepackage{tablefootnote}
\usepackage{makecell}
\usepackage{verbatim}
\usepackage{setspace}
\usepackage{floatrow}
\usepackage{adjustbox}
\usepackage{cleveref}
\usepackage{pifont}
\usepackage{longtable}
\usepackage{dashrule}
\usepackage{arydshln}

\usepackage{pgfplots}
\usepackage{times}
 
\usetikzlibrary{arrows.meta, positioning, fit, shapes.geometric, arrows}

\usepackage{ltxtable}

\usepackage{titlesec}
\titlespacing*{\paragraph}{0pt}{0.5em}{0.5em}  

\usepackage{algorithm}
\usepackage[%
    noEnd=true,%
    indLines=true,%
    italicComments=false,%
    commentColor=darkgreen,%
]{algpseudocodex}

\usepackage{wrapfig}

\theoremstyle{definition}

\theoremstyle{remark}

\makeatletter
\newcommand{\verbatimfont}[1]{\def\verbatim@font{#1}}%
\makeatother

\makeatletter
\patchcmd{\l@section}
  {\addvspace{1.0em \@plus\p@}}
  {}
  {}{}
\makeatother

\newcommand{\cuemail}[1]{\href{#1@cornell.edu}{#1}}

\renewcommand{\toprule}{\noalign{\hrule height 0.9pt\vspace{0.15cm}}}
\renewcommand{\midrule}{\noalign{\vspace{0.1cm}\hrule height 0.7pt\vspace{0.1cm}}}
\renewcommand{\bottomrule}{\noalign{\vspace{0.1cm}\hrule height 0.9pt\vspace{0.1cm}}}

\makeatletter
\DeclareDocumentCommand{\R}{O{} O{}}{
    \ifx&#1&\@empty\relax {
        \ifx&#2&\@empty\relax {
            \mathbb{R}
        } \else {
            \mathbb{R}^{#2}
        }
        \fi
    } \else {
        \ifx&#2&\@empty\relax {
            \mathbb{R}^{#1}
        } \else {
            \mathbb{R}^{#1 \times #2}
        }
        \fi
    } 
    \fi
}
\makeatother

\newcommand{\discrete}[1]{
    \vbox{%
        \hrule height 0.55pt  
        \kern0.3ex  
        \hbox{%
            \kern-0.0em  
            \ifmmode#1\else\ensuremath{#1}\fi  
            \kern-0.0em  
        }
    }
}

\DeclareSymbolFont{matha}{OML}{txmi}{m}{it}
\DeclareMathSymbol{\varv}{\mathord}{matha}{118}

\let\svthefootnote\thefootnote
\newcommand\freefootnote[1]{%
  \let\thefootnote\relax%
  \footnotetext{#1}%
  \let\thefootnote\svthefootnote%
}

\algnewcommand{\LeftComment}[1]{~~~~\textcolor{darkgreen}{\(\triangleright\) #1}}

\newfloatcommand{capbtabbox}{table}[][\FBwidth]

\title{Beyond Decoder-only: Large Language Models Can be Good Encoders for Machine Translation}

\author{Yingfeng Luo, Tong Zheng, Yongyu Mu, Bei Li, Qinghong Zhang, Yongqi Gao \\ [2pt]
    \textbf{Ziqiang Xu, Peinan Feng,  Xiaoqian Liu, Tong Xiao\thanks{Corresponding author}, Jingbo Zhu}  \\ [2pt]
    NLP Lab, Northeastern University, Shenyang, China \\ [2pt]
    NiuTrans Research, Shenyang, China \\ 
    \texttt{luoyingfeng\_neu@outlook.com} \\
    \texttt{\{\cuemail{xiaotong,zhujingbo}\}@mail.neu.edu.cn}
    \ifcolmpreprint%
        \\[3pt]
        (Version~$2$)
    \fi
}

\begin{document}
\maketitle

\addtocontents{toc}{\protect\setcounter{tocdepth}{-10}}
\begin{abstract}
    The field of neural machine translation (NMT) has changed with the advent of large language models (LLMs). 
Much of the recent emphasis in natural language processing (NLP) has been on modeling machine translation and many other problems using a single pre-trained Transformer decoder, while encoder-decoder architectures, which were the standard in earlier NMT models, have received relatively less attention. 
In this paper, we explore translation models that are universal, efficient, and easy to optimize, by marrying the world of LLMs with the world of NMT. 
We apply LLMs to NMT encoding and leave the NMT decoder unchanged. 
We also develop methods for adapting LLMs to work better with the NMT decoder. 
Furthermore, we construct a new dataset involving multiple tasks to assess how well the machine translation system generalizes across various tasks. 
Evaluations on the WMT and our datasets show that results using our method match or surpass a range of baselines in terms of translation quality, but achieve $2.4 \sim 6.5 \times$ inference speedups and a $75\%$ reduction in the memory footprint of the KV cache. 
It also demonstrates strong generalization across a variety of translation-related tasks.

\vspace{.3em}
\hspace{.5em}
\includegraphics[height=1.5em]{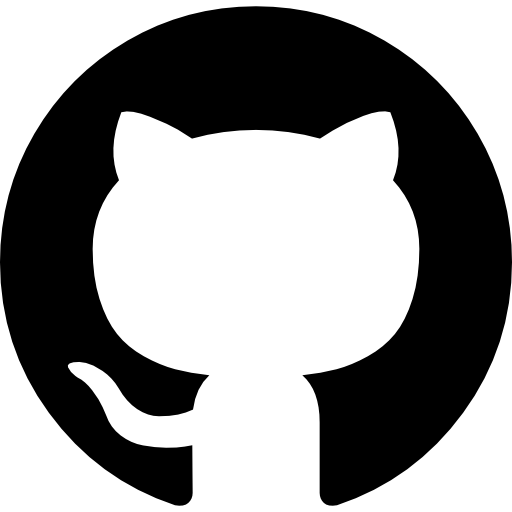}{\hspace{.75em}\parbox{\dimexpr\linewidth-2\fboxsep-2\fboxrule}{\vspace{-5pt} \href{https://github.com/NiuTrans/LaMaTE/}{NiuTrans/LaMaTE}}}

\end{abstract}
\section{Introduction}

\begin{figure}[t]
    \includegraphics[width=0.90\linewidth]{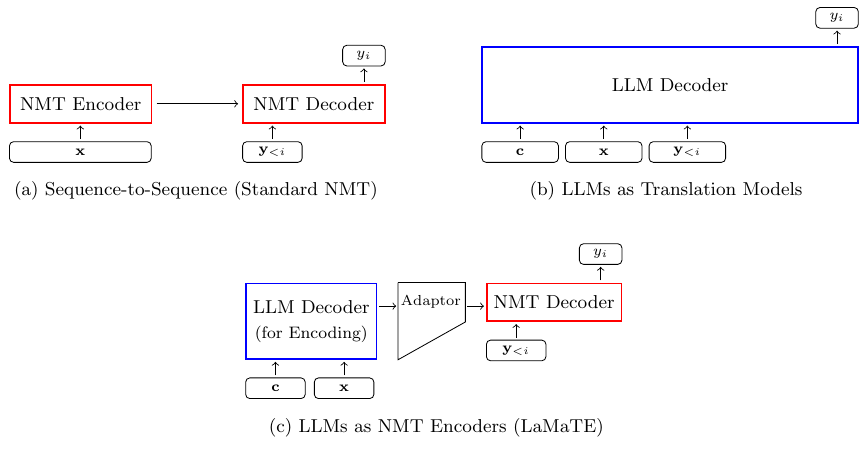}
    \vspace{-0.2cm}
    \caption{Architectures of machine translation models. In standard NMT models, an encoder is used to encode the source-language sequence x, and a decoder is used to generate the target-language sequence y from left to right. In LLMs, the decoder-only architecture is adopted. Both x and y, along with the prompt c, are represented as a single sequence, which is processed by a large decoding network. In the LaMaTE model, an LLM serves as the encoder. The output of the LLM is transformed into the input to the NMT decoder through an adaptor. The NMT decoder then generates the target-language sequence as usual.}
    \label{fig:mt-architectures}
\end{figure}

The last decade has yielded remarkable breakthroughs in machine translation (MT) through the use of deep neural networks, scaled dramatically in both model parameters and training data. 
During this period, early methods, known as neural machine translation (NMT), were largely based on the encoder-decoder architecture \citep{DBLP:conf/nips/SutskeverVL14,DBLP:journals/corr/BahdanauCB14}.  
In NMT, the machine translation problem is commonly treated as a sequence-to-sequence task, where the input sequence is first encoded into an intermediate representation, and the output sequence is then generated based on this representation, as illustrated in \cref{fig:mt-architectures} (a). 
Models of this kind are typically trained on bilingual text in a supervised manner, and their inference on modern GPUs is efficient. 
However, like their predecessors in past decades (e.g., statistical machine translation models), NMT models are generally developed for specific tasks, such as translation in a specific genre or domain.

Everything changed in NLP, with the success of large language models (LLMs) through large-scale self-supervised learning \citep{DBLP:conf/nips/BrownMRSKDNSSAA20}. 
In the LLM paradigm, the translation problem is framed as the token prediction problem in language modeling, as illustrated in \cref{fig:mt-architectures} (b). 
Such an approach greatly simplifies the modeling. We can now pre-train the LLM on large amounts of text via self-supervision so as to produce a single model that can be fine-tuned and prompted for various translation-related tasks, such as constrained translation and post-editing.  
Nevertheless, as a consequence of using large-scale neural networks, LLMs are computationally expensive and pose challenges for applications requiring low latency and a small memory footprint.

The fundamental change brought by LLMs has led most of the field to focus on replacing previous NLP systems with LLMs. 
However, machine translation researchers might still expect that a system could be efficient and easy to optimize, and, at the same time, could generalize across various sorts of tasks. 


In this paper, we examine how the world of LLMs can be married with the world of NMT, thereby benefiting from both paradigms. 
One simple approach is to use an LLM as the encoder in NMT (call it \textbf{LaMaTE} --- \textbf{L}arge L\textbf{a}nguage \textbf{M}odels as M\textbf{a}chine \textbf{T}ranslation \textbf{E}ncoders).  
As both the encoder and decoder are essentially based on language models, we can view the whole system as a single language model. 
The decoupling of encoding and decoding from language modeling confers great modeling flexibility. 
A direct result of this is that we can employ a deep, complex encoder and a lightweight decoder, as illustrated in \cref{fig:mt-architectures} (c). 
This heterogeneous architecture is particularly well-suited for machine translation, where we can use a powerful encoder to understand the input text, while generating high-quality translations at a lower decoding cost.

In order to examine to what extent the model can generalize, we develop a new benchmark, called the \textbf{Com}prehensive \textbf{M}achine \textbf{T}ranslation benchmark (\textbf{ComMT}). 
It consists of several diverse tasks that evaluate different aspects of a translation model. While there have been datasets developed for fine-tuning LLMs for multiple translation tasks \citep{DBLP:journals/corr/abs-2402-17733}, our focus is on a broader range of application scenarios. 
We hope that such a benchmark can be adopted to provide a systematic evaluation of machine translation systems, and that this, in turn, will encourage practitioners to pay more attention to the issue of generalization when developing these systems.

We conduct extensive experiments and evaluate various models including NMT, LLMs, and our LaMaTE method. 
Our results show that the LaMaTE model achieves comparable or better performance than a range of baseline systems on several tasks, but runs $2.4 \sim 6.5$ times faster and reduces the memory footprint of the KV cache by $75\%$. 
Evaluations on the ComMT dataset also demonstrate the strong generalization capabilities of the LaMaTE model, showing significant improvements over baseline systems. 
These results are not surprising, but intriguing, as findings in NMT remain applicable in the era of LLMs, for instance, scaling up the encoding network is still beneficial for machine translation tasks. 
This suggests an interesting direction for future work, where we could develop a powerful yet efficient system by using a strong model for language understanding and a lightweight model for language generation.

\section{Related Work}
\label{sec:model}

Designing widely applicable models for MT has been an active area of NLP research for decades. Although MT models have evolved significantly over time, most of them still operate on an  ``analyze-then-generate'' paradigm \citep{brown1993mathematics,koehn2003statistical,DBLP:journals/corr/BahdanauCB14}. For example, in statistical MT, the source-language sentence is parsed into either syntactic or non-syntactic forms, and the translation is generated by mapping these parsed forms to target-language constructions \citep{chiang2005hierarchical}. LLMs can broadly be categorized as following a similar design: they first compute the key-value cache for each input sequence, and then produce output tokens in a left-to-right manner based on this cache. From a modeling perspective, therefore, decoupling the encoding and decoding processes is a natural design choice for both traditional MT and LLMs.

In NLP, a large body of work has focused on pre-training and applying text encoders, such as the BERT series of models \citep{DBLP:conf/naacl/DevlinCLT19}, which are primarily designed to address language understanding problems. More recently, there have been attempts to use LLMs as text encoders, though these models are more commonly used for generating text. For example, \citet{DBLP:journals/corr/abs-2404-05961} and \citet{DBLP:journals/corr/abs-2402-09906} fine-tuned LLMs for text encoding and showed that LLMs can produce high-quality representations of text in various embedding tasks. However, it is rare to see studies on incorporating LLMs in the encoding for NMT or other language generation tasks.

In fact, the focus of research in MT has shifted. Recent studies in this field are now more concerned with the adaptation of LLMs, rather than the improvement of model architectures \citep{DBLP:conf/nips/LiZ0LGGTXZWC24,DBLP:conf/acl/ZhengLBXZ24}. 
One strand of research aims to enable LLMs to translate via prompting techniques, including designing better prompts \citep{DBLP:conf/icml/0006HB23,DBLP:conf/naacl/ZhuLDXHKCL24}, introducing translation demonstrations \citep{DBLP:conf/acl/MuRCFLLXZZ23,DBLP:conf/acl/ChitaleGD24}, and using chain-of-thought reasoning \citep{DBLP:conf/emnlp/LuYHZLW24}. 
Another strand focuses on developing MT-specific LLMs via fine-tuning \citep{DBLP:journals/corr/abs-2309-11674,DBLP:journals/corr/abs-2305-18098,DBLP:journals/corr/abs-2402-17733,DBLP:journals/corr/abs-2403-11430,zheng2025asymmetric}; 
we have followed this approach, but with a focus on improving the model architecture, as well as developing a new benchmark for evaluating universal MT models.

While LLMs have significantly enhanced MT with their superior versatility and generalization capabilities, much of the current research continues to focus on single-task contexts, which do not fully utilize the potential of LLMs to address diverse translation challenges.
The research highlighted in \citet{DBLP:journals/corr/abs-2402-17733} seeks to develop a more universal model by integrating a mix of general and translation-related tasks. 
Yet, despite these advancements, there remains a notable gap in the availability of comprehensive and high-quality data resources specifically tailored to the development of universal translation models.
Addressing this deficiency would maximize LLM capabilities and broaden their applicability across a variety of translation-related tasks.

This work is also related to efficient methods for LLMs. For example, one can compress the model using quantization and pruning techniques \citep{DBLP:conf/icml/XiaoLSWDH23,DBLP:conf/nips/MaFW23}, and speed-up inference using speculative decoding algorithms \citep{DBLP:conf/icml/LeviathanKM23,DBLP:conf/nips/KimMMMMGK23,hutowards}. But these methods and ours are in no way contradictory. Since our model follows standard encoding (or prefilling) and decoding frameworks, it can be easily combined with various efficient methods to further improve efficiency. 
\section{LaMaTE}
\label{sec:methodology}

In this section, we introduce LaMaTE and its training method.

\subsection{Model Architecture}

\begin{figure}[htbp]
    \includegraphics[width=\linewidth]{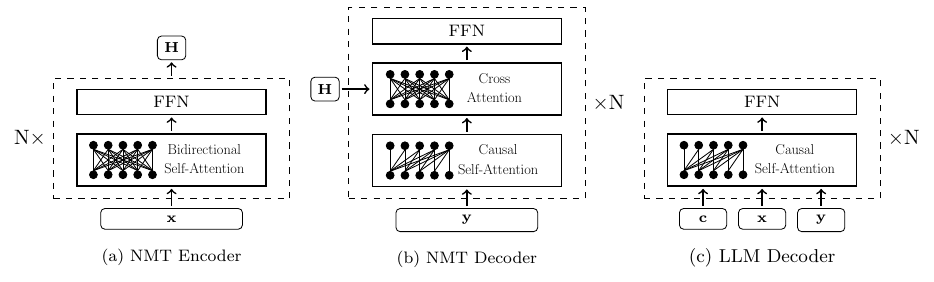}
    \vspace{-0.4cm}
    \caption{Architecture of the NMT Encoder, NMT Decoder, and LLM Decoder. We omit the layer normalization and residual connections for simplicity.}
    \label{fig:encoder_decoder_llm}
\end{figure}

We begin by outlining the basic concepts and notation needed for our description. 
There are three networks that we consider here, as illustrated in \cref{fig:encoder_decoder_llm}.

\begin{itemize}
\item NMT Encoder $\mathrm{Enc}(\mathbf{x})$. 
It is a standard Transformer encoder, consisting of an embedding layer and a number of stacked Transformer layers. 
The output of the NMT encoder is a bidirectional representation of the input token sequence $\mathbf{x}$, denoted by $\mathbf{H}$\footnote{$\mathbf{H}$ is a sequence of vectors, each corresponding to the representation at each position of $\mathbf{x}$.}.

\item NMT Decoder $\mathrm{Dec}(\mathbf{H},\mathbf{y})$. 
It shares the same architecture as the NMT encoder, with an additional cross-attention sublayer added in each Transformer layer. 
The NMT decoder accepts two inputs --- one from the encoder for cross-attention, and another as the regular sequential input. 
In the self-attention mechanism, each position can only attend to the preceding positions. 
So its output representation is unidirectional. 
Typically, a Softmax layer is added on top of the decoder to generate distributions of tokens.

\item LLM Decoder $\mathrm{Dec}(\underline{\hspace{1em}},[\mathbf{c},\mathbf{x},\mathbf{y}])$. 
It operates on token sequences only, without the need for input from the encoder. 
Thus, the cross-attention sublayers are removed, and the first parameter of the above function is left blank. 
Here $[\mathbf{c},\mathbf{x},\mathbf{y}]$ denotes the concatenation of the prompt $\mathbf{c}$, the input sequence $\mathbf{x}$, and the translation $\mathbf{y}$.
\end{itemize}

For NMT models, we can simply connect the NMT encoder and decoder together. 
The probability of token prediction can be expressed as
\begin{eqnarray}
\Pr(y_i|\mathbf{x},\mathbf{y}_{<i}) & = & \mathrm{Softmax}(\mathbf{W}\mathbf{S})_i \\
\mathbf{S} & = & \mathrm{Dec}(\mathbf{H},\mathbf{y}_{<i}) \\
\mathbf{H} & = & \mathrm{Enc}(\mathbf{x})
\end{eqnarray}

\noindent where $y_i$ denotes the target-language token at position $i$, $\mathbf{y}_{<i}$ denotes the target-language tokens that precede position $i$, and $\mathrm{Softmax}(\mathbf{W}\mathbf{S})_i$ denotes the Softmax function that computes the distribution of tokens at position $i$. 
$\mathbf{W}\mathbf{S}$ maps the decoder output $\mathbf{S}$ to a representation space of the vocabulary size using the linear mapping matrix $\mathbf{W}$.

The LaMaTE model follows this encode-decode architecture, but with an LLM decoder replacing the NMT encoder, given by
\begin{eqnarray}
\mathbf{S} & = & \mathrm{Dec}_{\phi}(\mathbf{H}',\mathbf{y}_{<i}) \label{eq:lamate-decoder} \\
\mathbf{H}' & = & \mathrm{Dec}_{\theta}(\underline{\hspace{1em}}, [\mathbf{c},\mathbf{x}]) \label{eq:lamate-encoder}
\end{eqnarray}

\noindent Here we use the subscripts $\phi$ and $\theta$ to denote the parameters for the NMT decoder and the LLM decoder, respectively. 

The function $\mathrm{Dec}_{\theta}(\underline{\hspace{1em}}, [\mathbf{c},\mathbf{x}])$ serves as a unidirectional encoder. 
Although it is generally thought that bidirectional encoders can make better use of contextual information, our experiments in \cref{sec:ablation} demonstrate that using such unidirectional encoders based on LLMs is also very effective. 
Given that both the encoder and decoder are unidirectional, we can roughly think of the whole model as a hybrid language model: a large, powerful model is used for prefilling, and a small, efficient model is used for decoding.

The use of pre-trained LLMs for encoding allows the model to have a stronger understanding of the input sequence, which has been found to be very beneficial in NMT systems \citep{DBLP:conf/emnlp/DouTWSZ18,DBLP:conf/acl/WangLXZLWC19}. 
As another bonus, one can easily adapt the model via fine-tuning and prompting, thereby deploying a single such system for many different tasks. 
For example, by giving an appropriate prompt $\mathbf{c}$,  we can guide the system to produce outputs following specific needs or contexts. 
This gives LaMaTE a big advantage over traditional NMT models.

One problem with the model described in Eqs. (\ref{eq:lamate-decoder}-\ref{eq:lamate-encoder}) is that the LLM decoder may have a larger hidden size than that of the NMT decoder. 
In this case, we add an adaptor to the output of the LLM decoder to reduce its dimensionality, so that it matches the NMT decoder. 
We can then redefine $\mathbf{H}'$ as
\begin{eqnarray}
\mathbf{H}' & = & F_{\omega}(\mathrm{Dec}_{\theta}(\underline{\hspace{1em}}, [\mathbf{c},\mathbf{x}]))
\end{eqnarray}

\noindent where $F_{\omega}(\cdot)$ is the adaptor with the parameters $\omega$.

\subsection{Adaptor}

\begin{figure}[t]
    \includegraphics[width=\linewidth]{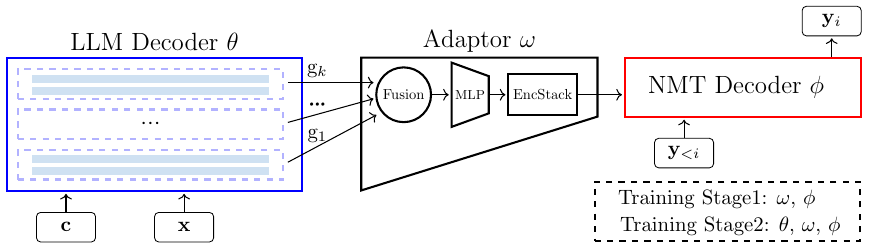}
    \caption{The architecture of LaMaTE, where the Adaptor consists of three components: Fusion combines the representations of layer groups $\mathbf{g}_k$, MLP reduces the representations's dimensionality, and EncStack learns bidirectional representations. The training process consists of two stages: the first stage trains the Adaptor and Decoder, and the second stage trains all model parameters.}
    \label{fig:lamate_detail}
\end{figure}

One simple form of adaptors is a linear mapping of the output of the LLM decoder, that is, we linearly transform representations from $\mathrm{Dec}_{\theta}(\cdot)$ to lower-dimensional representations required by $\mathrm{Dec}_{\phi}(\cdot)$. 
But we find that this method falls short of our expectations because LLMs are typically too big that simply transforming their output representations into lower-dimensional ones makes it difficult for untrained smaller models to use them directly.

Instead, we design an adaptor that makes better use of the LLM outputs for the NMT decoder. 
It involves three components.
The complete model architecture is illustrated in \cref{fig:lamate_detail}.

\begin{itemize}
\item Instead of considering only the output of the final layer of the LLM, we fuse the outputs of the intermediate layers to form a finer-grained output of the LLM decoder. 
Such methods have been found to be very beneficial to deep-to-shallow architectures in NMT \citep{DBLP:conf/acl/WangLXZLWC19,DBLP:journals/taslp/YangYYMHZWL23}. 
To do this, we divide the LLM layers into $K$ groups, and fuse their outputs in the following form
\begin{eqnarray}
\hspace{-0.5cm} \mathbf{H}_{\mathrm{fuse}} & = & \mathrm{LayerNorm} \left( \frac{1}{K} \sum_{k=1}^{K} w_k \mathbf{g}_k \right)
\end{eqnarray}

\noindent where $\mathrm{LayerNorm}(\cdot)$ is the layer normalization function, $\mathbf{g}_k$ is the output of the $k$-th layer group, and $w_k$ is the learnable weight for $\mathbf{g}_k$. We use the last layer's hidden states of each group as that group’s hidden states.

\item We place a 2-layer MLP after layer fusion to reduce the hidden size of $\mathbf{H}_{\mathrm{fuse}}$ (denoted by $d_1$) to the hidden size of the NMT decoder (denoted by $d_2$). 
This network is given by
\begin{eqnarray}
\mathbf{H}_{\mathrm{mlp}} & = & \mathrm{GELU}(\mathbf{H}_{\mathrm{fuse}} \mathbf{W}_1) \mathbf{W}_2
\end{eqnarray}

\noindent where $\mathrm{GELU}(\cdot)$ is the activation function, and $\mathbf{W}_1 \in \mathbb{R}^{d_1 \times d_2}$ and $\mathbf{W}_2 \in \mathbb{R}^{d_2 \times d_2}$ are the linear  mapping matrices.
\item It is also possible to learn bidirectional representations from unidirectional representations. A common method is to incorporate some Transformer encoder layers. Thus the self-attention models can help generate bidirectional representations. The final output of the adaptor is then given by
\begin{eqnarray}
\mathbf{H}' & = & \mathrm{EncStack}(\mathbf{H}_{\mathrm{mlp}})
\end{eqnarray}

\noindent where $\mathrm{EncStack}(\cdot)$ is a stack of Transformer encoder layers. Note that $\mathrm{EncStack}(\cdot)$ consists of only a few layers, and thus adds very small computational overhead. This step of bidirectional representation learning is optional, and one can choose whether to adopt it in practice.
\end{itemize}

\subsection{Model Training}

The LaMaTE model has three sets of parameters: $\theta$ for the LLM decoder, $\phi$ for the NMT decoder, and $\omega$ for the adaptor. Here, $\theta$ is initialized with the pre-trained parameters of the LLM, while $\phi$ and $\omega$ need to be trained from scratch.

Optimizing these parameters on bitext seems straightforward, but poses practical challenges. Since fine-tuning pre-trained LLMs is costly \citep{hu2022lora,zhang2025uorauniformorthogonalreinitialization}, using less labeled data is generally favorable. However, we found that this approach can lead to inadequate learning of the new parameters $\phi$ and $\omega$. 
On the other hand, extensive fine-tuning may cause the LLM to forget its original knowledge encoded in $\theta$. 
Similar findings have been reported in the literature \citep{DBLP:journals/corr/abs-2309-11674}.

Here, we present a two-stage training method that performs more efficient learning for different types of parameters. In the first stage, we freeze $\theta$ and pre-train both $\phi$ and $\omega$. The task of pre-training is a standard translation task. We train the model to translate source-language sequences to corresponding target-language sequences. Therefore, the adaptor and the NMT decoder can learn to map from source-language representations to translations. Since the parameters of the LLM decoder are frozen, this process requires only the forward pass of this large network, without the need for the backward pass. It is thus less computationally expensive than LLM fine-tuning, making it possible to scale up the pre-training to large datasets.

The second stage is a fine-tuning stage, where $\theta$, $\phi$, and $\omega$ are fine-tuned together on various tasks. 
For each task, as with instruction fine-tuning for LLMs, we provide LaMaTE with task-specific instructions, inputs and outputs, and optimize all the parameters end-to-end. 
In this way, the model is adapted to follow these instructions and can be deployed as a single, universal model to handle a variety of translation-related problems. We use our ComMT dataset for fine-tuning in this work.

\section{ComMT}

\subsection{Tasks}
Many translation-related tasks, such as document-level translation, rely on both understanding complex inputs and generating coherent, contextually appropriate outputs, which are not captured by commonly used sentence-level translation tasks. 
To generalize LaMaTE to diverse tasks and evaluate it on these tasks, we developed a new evaluation benchmark called ComMT. 
It comprises five tasks, with examples for each provided in \cref{tab:test example}:

\begin{itemize}
\item General Translation. This is a standard sentence in, sentence out task in general domains, serving as the foundation for more specialized tasks.
\item Document-level Translation. This task extends translation from sentence-level to document-level, focusing on maintaining coherence and context in extended texts, rather than merely achieving sentence-level accuracy. Our document-level translation tasks mainly involve general long texts, such as news articles.
\item Domain Translation. This task focuses on domain-specific translation, ensuring the accurate use of domain-specific terminology and expressions. Based on the application scenario and existing research, we identified five initial domains for study: Medical, Law,  Information Technology(IT), Colloquial, and Literature.
\item Terminology-constrained Translation. In this task, the system is required to produce translations that follow the given terminology translation requirements.
\item Automatic Post-editing. This task focuses on enhancing the quality of preliminary machine translation outputs by automatically correcting errors in grammar, spelling, and style in the initial translations.
\end{itemize}

We collected data mainly from public resources, driven by two goals: 
(i) to find as many data sources as possible in order to increase data \textbf{diversity}, and (ii) to collect as much \textbf{high-quality}, manually annotated translation data as possible. 
ComMT is multilingual and supports four languages: German, Czech, Russian, and Chinese. 
The reader can refer to Appendix \cref{sec:app_benchmark} for more process details of ComMT.

\begin{figure*}[t]
    \centering
    \includegraphics[width=0.98\linewidth]{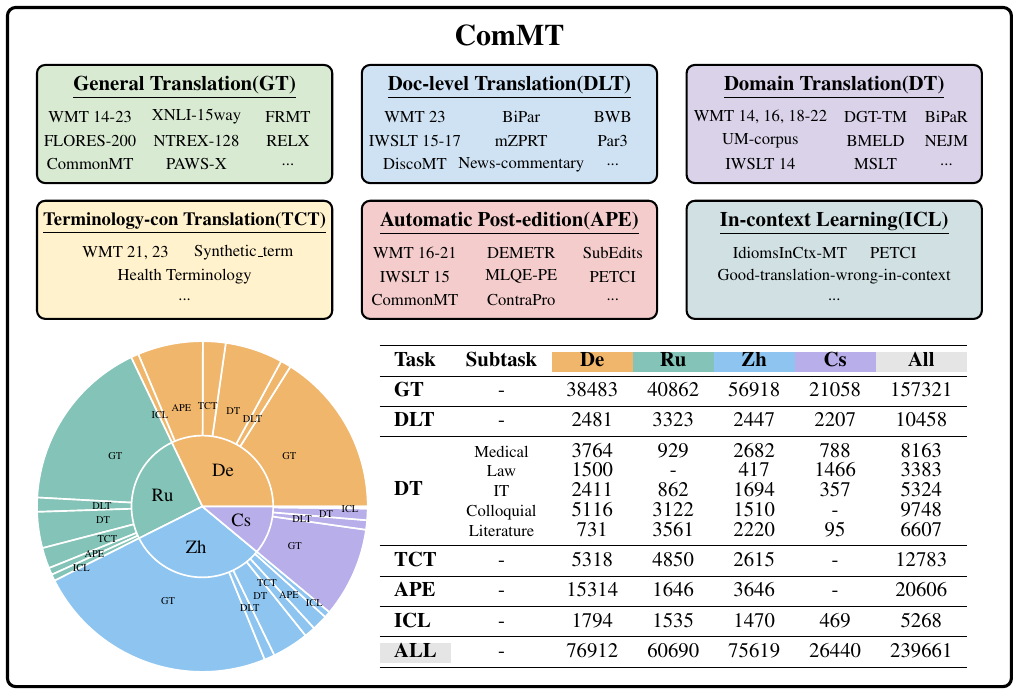}
    \caption{Our comprehensive translation dataset, \textbf{ComMT}, includes diverse translation-related tasks. The table presents the training set statistics for ComMT.}
    \label{fig.dataset}
\end{figure*}

Ultimately, we created a training set of 239k samples, as shown in \cref{fig.dataset}.
Additionally, we also created a test set, detailed in \cref{tab:benchmark_test}. 
For the General Translation task, we used established test sets, while for other tasks, we carefully curated test data by reorganizing existing datasets. 
We ensured that the source language in each case was original to avoid translationese issues \citep{DBLP:conf/emnlp/GrahamHK20,DBLP:journals/jair/LaubliCNSST20}.
However, some languages currently lack test sets because obtaining the task data is challenging. 
In the future, we plan to continue expanding this dataset to support more languages and tasks.

\renewcommand{\arraystretch}{2.5}
\begin{table*}[t]
  \centering
  \begin{adjustbox}{max width=\textwidth}
  \begin{tabular}{lp{5cm}cccccccc}
    \toprule
     \multirow{3}{*}{\Huge \textbf{Task}} &\multirow{3}{*}{\Huge \textbf{Subtask}} &\multicolumn{2}{c}{\Huge \textbf{De}} &\multicolumn{2}{c}{\Huge \textbf{Ru}} &\multicolumn{2}{c}{\Huge \textbf{Zh}} &\multicolumn{2}{c}{\Huge \textbf{Cs}} \\  %
      \cmidrule(lr){3-4} \cmidrule(lr){5-6} \cmidrule(lr){7-8} \cmidrule(lr){9-10} 
     &\Huge  &\Huge De2En &\Huge En2De &\Huge Ru2En &\Huge En2Ru &\Huge Zh2En &\Huge En2Zh &\Huge Cs2En &\Huge En2Cs \\
    \midrule
    \multirow{3}{*}{\Huge \textbf{General Translation}} &\Huge WMT23 &\Huge 549 &\Huge 557 &\Huge 1723 &\Huge 2074 &\Huge 1976 &\Huge 2074 &\Huge - &\Huge 2074 \\ 
     &\Huge  WMT22 &\Huge1984	&\Huge2037	&\Huge2016	&\Huge2037	&\Huge1875	&\Huge2037	&\Huge1448	&\Huge2037\\
     &\Huge FLORES-200 &\Huge - &\Huge 1012 &\Huge - &\Huge 1012 &\Huge - &\Huge 1012 &\Huge - &\Huge 1012 \\  \midrule
    \Huge \textbf{Doc-level translation} &\Huge \centering - &\Huge 500	&\Huge547	&\Huge-	&\Huge500	&\Huge415	&\Huge500	&\Huge500	&\Huge519 \\ \midrule
    \multirow{5}{*}{\Huge \textbf{Domain Translation}}  &\Huge Medical &\Huge 748 &\Huge 744 &\Huge 513 &\Huge 1772 &\Huge 580 &\Huge 2094 &\Huge 999 &\Huge -\\
     &\Huge Law &\Huge - &\Huge 969 &\Huge - &\Huge - &\Huge - &\Huge - &\Huge - &\Huge 959 \\
     &\Huge IT &\Huge - &\Huge - &\Huge - &\Huge 650 &\Huge - &\Huge - &\Huge - &\Huge 519 \\
     &\Huge Colloquial &\Huge 517 &\Huge 595	&\Huge 564	&\Huge 507	&\Huge 520	&\Huge 652 &\Huge - &\Huge - \\
     &\Huge Literature &\Huge 506	&\Huge-	&\Huge506	&\Huge-	&\Huge401	&\Huge550 &\Huge - &\Huge - \\
      \midrule
    \Huge \textbf{Terminology-constrained Translation} &\Huge \centering - &\Huge 2948 &\Huge 775 &\Huge - &\Huge 500 &\Huge 1638 &\Huge 798 &\Huge - &\Huge 2898 \\ \midrule
    \Huge \textbf{Automatic Post-edition} &\Huge \centering - &\Huge 1980 &\Huge 1000 &\Huge 526 &\Huge 1023 &\Huge - &\Huge 1000 &\Huge - &\Huge - \\ 
    \bottomrule
  \end{tabular}
  \end{adjustbox}
  \caption{Statistics for the ComMT Test Set. We ensure that the source language of the test set is original, so not all language directions and tasks have corresponding data due to the difficulty in collecting certain datasets.}
  \label{tab:benchmark_test}
\end{table*}
\renewcommand{\arraystretch}{1}

\subsection{Quality Verification}
\label{sec:data_verification}
\begin{figure*}[t]
    \centering
    \includegraphics[width=0.95\linewidth]{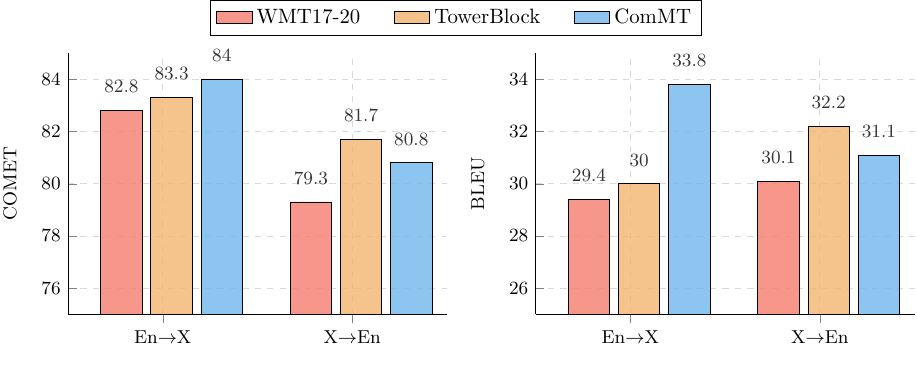}
    \caption{Comparison of performance across three datasets---WMT17-20, TowerBlock, and ComMT---fine-tuned on Llama3-8B and evaluated on the WMT23 test set.}
    \label{fig:compared_commt}
\end{figure*}

To verify the quality of our dataset, we fine-tuned the Llama3-8B model \citep{DBLP:journals/corr/abs-2407-21783} and compared it against two other datasets commonly used in LLM-based MT research: (i) a merged news test set from WMT17 to WMT20 with 61k samples, widely adopted data setting in previous studies \citep{DBLP:journals/corr/abs-2309-11674,DBLP:conf/emnlp/Jiao0W0LW0T23,DBLP:journals/corr/abs-2403-11430}; (ii) the TowerBlock dataset \citep{DBLP:journals/corr/abs-2402-17733}, which comprises 638k samples across translation tasks, named entity recognition (NER), general dialogue, and other tasks.
We evaluated the resulting models on the WMT23 test set, with the results shown in \cref{fig:compared_commt}. 

The results clearly demonstrate that ComMT surpasses WMT17–WMT20 in both the En $\to$ X and X $\to$ En directions. 
However, while it excels in the En $\to$ X direction, it underperforms in the X $\to$ En direction compared to TowerBlock.
This discrepancy may be attributed to TowerBlock’s larger dataset size and its inclusion of additional general task data, predominantly in English, which may specifically enhance performance when translating into English.
We intend to investigate this further in future research.

Overall, ComMT exhibits significant diversity and broad applicability, providing a well-curated data resource for developing and evaluating universal translation models.

\section{Experimental Setup}
\subsection{Data and Evaluation Metrics}
We trained our model on multilingual translations between English (En) and four languages: German (De), Czech (Cs), Russian (Ru), and Chinese (Zh), resulting in a total of eight translation directions. 
The parallel corpus for training stage 1 was sourced from WMT2023, as detailed in \cref{tab:parallel_data}. 
We sampled 10M bilingual sentence pairs for each language, yielding a total of 40M pairs for training stage 1.
The training stage 2 utilizes data from the ComMT training set with 239k samples.

We conducted tests on ComMT and evaluated the model's translation performance across all tasks using COMET (wmt22-comet-da) \citep{DBLP:conf/emnlp/ReiSFL20} and SacreBLEU \citep{DBLP:conf/wmt/Post18}. 
Additionally, we used Terminology Success Rate (TSR) for the terminology-constrained translation task and Human Translation Edit Rate (HTER) for the automatic post-editing task.

\subsection{Training}

\begin{figure*}[!ht]
    \centering
    \includegraphics[width=1.0\linewidth]{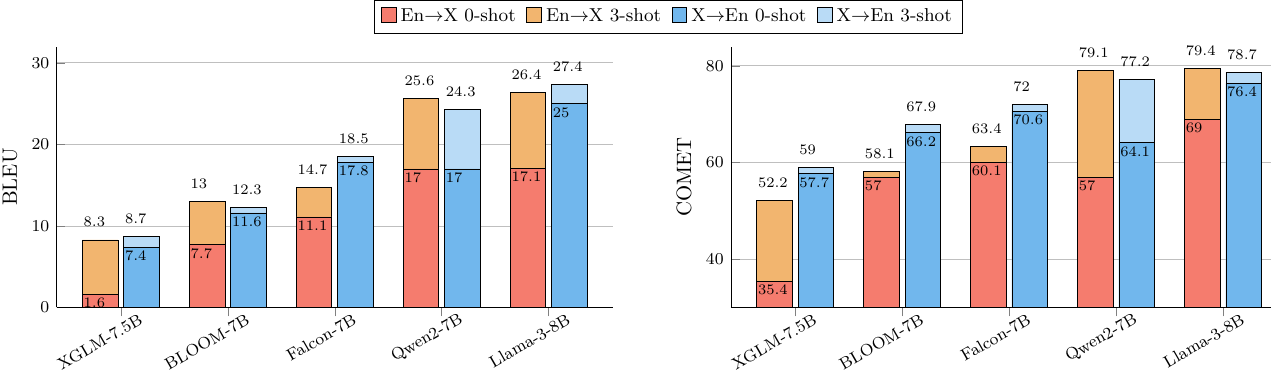}
    \caption{Evaluation of LLM translation capabilities in 0-shot and 3-shot settings using the WMT23 test set.}
    \label{fig.shot_result}
\end{figure*}

Our multilingual translation experiments necessitate a language model with multilingual capabilities for better encoding.
To assess the fundamental translation capabilities of state-of-the-art multilingual LLMs, we conducted a comprehensive evaluation across several widely used models, including XGLM-7.5B \citep{DBLP:conf/emnlp/LinMAWCSOGBDPSK22}, Bloom-7B \citep{DBLP:journals/corr/abs-2211-05100}, Falcon-7B \citep{DBLP:journals/corr/abs-2311-16867}, Qwen2-7B \citep{DBLP:journals/corr/abs-2407-10671}, Llama3-8B \citep{DBLP:journals/corr/abs-2407-21783}.
These models were evaluated in both zero-shot and three-shot settings across English ~$\leftrightarrow$~ German, Czech, Russian, and Chinese translation tasks.
The averaged performance in in \cref{fig.shot_result} show that Llama3-8B consistently outperformed the other models in both En $\to$ X and X $\to$ En translation directions.
Its superior performance across different language pairs suggests that it possesses stronger multilingual capabilities.
Based on these results, we selected Llama3-8B as our backbone model for subsequent experiments.

Llama3-8B has a model dimension of 4096 and consists of 32 layers. For the decoder, we set the dimension to 1024 with 8 layers, which also applies to the EncStack of the adaptor, with the adaptor and decoder together introducing fewer than 500M parameters.
Hyperparameters for training are detailed in \cref{tab:hyperparameters}.

\subsection{Models for Comparison}
We create two categories of models for comparison:

\paragraph{Decoder-only Models.} 
This includes: BigTranslate \citep{DBLP:journals/corr/abs-2305-18098}, Bayling-13B \citep{DBLP:journals/corr/abs-2306-10968}, Aya-23-8B \citep{DBLP:conf/acl/UstunAYKDOBSOKV24}, ALMA-7B \citep{DBLP:journals/corr/abs-2309-11674} and TowerInstruct-7B \citep{DBLP:journals/corr/abs-2402-17733}. Also included are variants of Llama3-8B: Llama3-8B-Base (3-shot in-context learning), Llama3-8B-Inst (general instruction-tuned), and Llama3-8B-SFT (fine-tuned on ComMT).

\paragraph{Encoder-Decoder Models.} 
This includes:  
(i) NLLB-3.3B \citep{DBLP:journals/corr/abs-2207-04672}: a comprehensive multilingual model.
(ii) Traditional encoder-decoder models: 
NMT-40-8 (40 encoder layers, 8 decoder layers) and NMT-8-8 (8 encoder layers, 8 decoder layers), both with a model dimension of 1024, trained from scratch on 161M bilingual sentences from combined parallel data of four languages.
(iii) Fine-tuned encoder-decoder models:  mT5-Large (1.2B) \citep{DBLP:conf/naacl/XueCRKASBR21}, aligned with our method's data settings.


\section{Results and Analyses}

\subsection{Main Results}

\newcommand{\myHuge}{\fontsize{30}{30}\selectfont}
\newcommand{\myhuge}{\fontsize{25}{25}\selectfont}
\setlength\dashlinedash{8pt}
\setlength\dashlinegap{8pt}
\renewcommand{\arraystretch}{3.5}
\begin{table*}[htbp]
  \tabcolsep=10pt
  \centering
  \begin{adjustbox}{max width=\textwidth}
  \begin{tabular}{lcccccccccccc}
    \toprule
     \multirow{2}{*}{\textbf{\myHuge Model}} &\multicolumn{2}{c}{\textbf{\Huge General Trans}} &\multicolumn{2}{c}{\textbf{\Huge Domain Trans}} &\multicolumn{2}{c}{\textbf{\Huge Doc-level Trans}} &\multicolumn{3}{c}{\textbf{\Huge Terminology-con Trans}} &\multicolumn{3}{c}{\textbf{\Huge Automatic Post-edition}} 
     \vspace{0em}\\
     \cmidrule(lr){2-3} \cmidrule(lr){4-5} \cmidrule(lr){6-7} \cmidrule(lr){8-10} \cmidrule(lr){11-13} 
    &\textbf{\myhuge COMET} &\textbf{\myhuge BLEU} &\textbf{\myhuge COMET} &\textbf{\myhuge BLEU} &\textbf{\myhuge COMET} &\textbf{\myhuge BLEU} &\textbf{\myhuge COMET} &\textbf{\myhuge BLEU} &\textbf{\myhuge TSR $\uparrow$} &\textbf{\myhuge COMET} &\textbf{\myhuge BLEU} &\textbf{\myhuge HTER $\downarrow$}\\
    \midrule
    \multicolumn{13}{c}{\textbf{\textit{\myHuge Decoder-only Models}}}\\
     
     \myHuge ALMA-7B* &\myHuge \textbf{83.23} &\myHuge 31.49 &\myHuge 83.93 &\myHuge 31.94 &\myHuge 82.03 &\myHuge 18.59 &\myHuge 81.84 &\myHuge 24.89 &\myHuge 43.20 &\myHuge 85.25 &\myHuge 40.91 &\myHuge 54.42  \\
     \myHuge BigTranslate* &\myHuge 77.84 &\myHuge 22.04 &\myHuge 77.53 &\myHuge 21.99 &\myHuge 72.36 &\myHuge 9.18 &\myHuge 77.29 &\myHuge 18.56 &\myHuge 33.10 &\myHuge 78.84 &\myHuge 29.82 &\myHuge 69.04 \\
     \myHuge Aya-23-8B  &\myHuge 82.72 &\myHuge 32.11 &\myHuge 84.07 &\myHuge 33.67 &\myHuge \textbf{84.98} &\myHuge \textbf{32.19} &\myHuge 81.47 &\myHuge 31.14 &\myHuge 74.15 &\myHuge 87.68 &\myHuge 67.92 &\myHuge 36.23 \\
     \myHuge TowerInstruct-7B  &\myHuge \underline{83.12} &\myHuge \underline{32.75} &\myHuge 83.65 &\myHuge 33.25 &\myHuge 81.61 &\myHuge 28.22 &\myHuge 82.57 &\myHuge 31.84 &\myHuge 76.98 &\myHuge \textbf{88.91} &\myHuge \textbf{77.41} &\myHuge \textbf{23.49} \\
     \myHuge Bayling-13B &\myHuge 74.81 &\myHuge 21.34 &\myHuge 76.50 &\myHuge 24.07 &\myHuge 74.10 &\myHuge 19.46 &\myHuge 77.01 &\myHuge 25.08 &\myHuge 71.94 &\myHuge 73.90 &\myHuge 29.67 &\myHuge 114.80 \\
      \myHuge Llama3-8B-Base &\myHuge 79.03 &\myHuge 26.89 &\myHuge 81.21 &\myHuge 28.29 &\myHuge 80.44 &\myHuge 26.77 &\myHuge 81.81 &\myHuge 29.84 &\myHuge 79.20 &\myHuge 86.26 &\myHuge 61.43 &\myHuge 38.93 \\
     \myHuge Llama3-8B-Inst &\myHuge 73.56 &\myHuge 23.20 &\myHuge 73.04 &\myHuge 22.69 &\myHuge 81.79 &\myHuge 27.22 &\myHuge 71.36 &\myHuge 24.41 &\myHuge \underline{86.46} &\myHuge 68.05 &\myHuge 25.70 &\myHuge 101.85 \\
     \myHuge Llama3-8B-SFT &\myHuge 82.41 &\myHuge 32.42 &\myHuge \underline{84.60} &\myHuge \underline{36.92} &\myHuge 83.79 &\myHuge \underline{31.75} &\myHuge \textbf{84.01} &\myHuge \textbf{36.77} &\myHuge \textbf{87.52} &\myHuge 88.57 &\myHuge \underline{70.13} &\myHuge \textbf{31.57} \\
    \midrule
    
    \multicolumn{13}{c}{\textbf{\textit{\myHuge Encoder-Decoder Models}}}\\
    \myHuge NLLB-3.3B*  &\myHuge 81.30  &\myHuge 31.63 &\myHuge 81.38 &\myHuge 32.66 &\myHuge 73.60 &\myHuge 11.72 &\myHuge 80.21 &\myHuge 26.32 &\myHuge 41.25 &\myHuge 84.97 &\myHuge 46.42 &\myHuge 52.42 \\
     \myHuge NMT-8-8* &\myHuge 79.70 &\myHuge 30.08 &\myHuge 80.63 &\myHuge 32.76 &\myHuge 73.98 &\myHuge 10.11 &\myHuge 77.03 &\myHuge 25.54 &\myHuge 40.56 &\myHuge 84.50 &\myHuge 47.52 &\myHuge 50.21\\
     \myHuge NMT-40-8* &\myHuge 80.89 &\myHuge 32.05 &\myHuge 81.50 &\myHuge 34.06 &\myHuge 74.74 &\myHuge 12.39 &\myHuge 77.19 &\myHuge 25.80 &\myHuge 41.15 &\myHuge 85.10 &\myHuge 47.57 &\myHuge 50.26 \\
     \myHuge mT5-Large &\myHuge 81.26 &\myHuge 29.34 &\myHuge 82.06 &\myHuge 30.26 &\myHuge 77.69 &\myHuge 14.43 &\myHuge 81.80 &\myHuge 30.34 &\myHuge 76.93 &\myHuge 85.88 &\myHuge 62.77 &\myHuge 45.10 \\
    \hdashline
    
     \myHuge LaMaTE (Ours) &\myHuge 82.32 &\myHuge \textbf{33.85} &\myHuge \textbf{84.69} &\myHuge \textbf{37.49} &\myHuge \underline{84.34} &\myHuge 31.69 &\myHuge \underline{83.35} &\myHuge \underline{34.76} &\myHuge 75.64 &\myHuge \underline{88.60} &\myHuge 69.10 &\myHuge 33.42 \\
    \bottomrule
  \end{tabular}
  \end{adjustbox}
  \caption{Performance on ComMT test set. \textbf{Bold} numbers represent the highest scores in each category, while \underline{underlined} numbers indicate the second highest scores. Models marked with "*" cannot handle additional inputs for the terminology-constrained translation and automatic post-editing tasks, we only use the source sequence as input.}
  \label{tab:performance} 
\end{table*}
\renewcommand{\arraystretch}{1}

\begin{figure*}[b]
    \centering
    \includegraphics[width=1.0\linewidth]{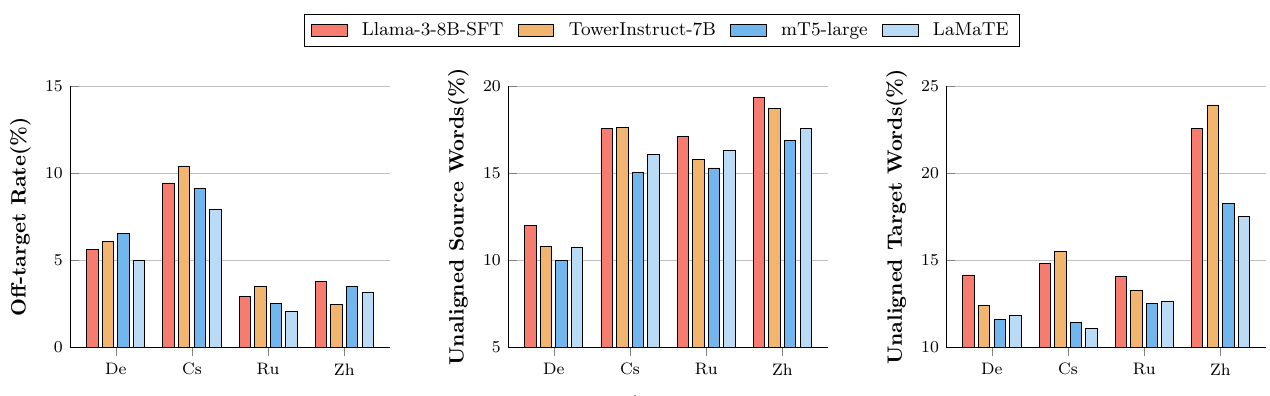}
    \caption{Comparison of decoder-only (Llama3-8B and TowerInstruct-7B) and encoder-decoder (mT5-large and LaMaTE) models on off-target rate (OTR), unaligned source words (USW), and unaligned target words (UTW). }
    \label{fig:misalignment}
\end{figure*}

The averaged performance across all directions is shown in \cref{tab:performance}, with more detailed results provided in \cref{tab:detail_performance}.

\paragraph{Comparison with Encoder-Decoder Models.} In comparison to encoder-decoder models, LaMaTE demonstrates stronger performance, surpassing NLLB-3.3B, NMT, and mT5-Large across all evaluated tasks. 
This result hightlights the potential of LLMs as high-capacity encoders, capable of producing more expressive and informative representations. These enhanced representations provide a solid foundation for the decoder, ultimately improving translation quality.
Moreover, LaMaTE's consistent performance across multiple tasks underscores its strong generalization capability.

\paragraph{Comparison with Fine-Tuned LLMs.} When compared with fine-tuned LLMs, LaMaTE achieves comparable overall performance to Llama3-8B-SFT. 
While Llama3-8B-SFT holds a slight advantage in the terminology-constrained translation task with higher TSR scores, there is no significant difference in other tasks. 
Additionally, LaMaTE outperforms earlier fine-tuned LLMs like Bayling-13B and remains competitive with more advanced models such as TowerInstruct-7B. 
This further demonstrates that LLMs can serve as effective MT encoders, offering an effective and computationally efficient alternative to the approach of fine-tuning LLMs as direct generators.

\paragraph{Misalignment Evaluation.} To further investigate translation quality, we analyze misalignment issues commonly observed in LLM-based generation \citep{DBLP:conf/acl/ZhangC0X024,DBLP:conf/acl/ZengMYZ24}, where models may generate output that diverges from the source text.
We compare decoder-only models (Llama3-8B and TowerInstruct-7B) and encoder-decoder (mT5-large and LaMaTE) across three metrics: off-target rate, unaligned source words, and unaligned target words in En $\to$ X direction. 
Refer to the calculation details in \cref{sec:result_app}.
As shown in \cref{fig:misalignment}, encoder-decoder models consistently exhibit lower scores across all three dimensions, indicating they reduce misalignment compared to decoder-only models.
LaMaTE follows the encoder-decoder paradigm, which also helps mitigate misalignment issues as an added bonus.
We speculate that this improvement can be attributed to the cross-attention mechanism in the NMT decoder, which allows direct token-wise interaction between the target and the source sequence, thereby enhancing source-target alignment.
In contrast, LLMs rely on a single attention mechanism to jointly process the concatenated source and target text, which could potentially cause a loss of attention focus \citep{DBLP:conf/acl/ZhangC0X024}.

\begin{table}[t]
  \centering
  \begin{adjustbox}{max width=\textwidth}
  \begin{tabular}{lcccc}
    \toprule
     \textbf{Models} & \textbf{HRL(12)} &\textbf{MRL(23)} &\textbf{LRL(22)} & \textbf{XRL(42)} \\
    \midrule
    NMT-40-8  &30.13 &30.28 &21.02 &16.59 \\
    mT5-large &29.12 &30.14 &\textbf{27.46} &\textbf{24.64}  \\
    \hdashline
    LaMaTE (Ours) &\textbf{32.11} &\textbf{32.13} &26.20 &21.70  \\
    \bottomrule
  \end{tabular}
  \end{adjustbox}
  \caption{BLEU performance evaluated on the FLORES-200 devtest set, with models trained on OPUS-100 data  (99 languages). HRL, MRL, LRL, and XRL represent high, medium, low, and very low-resource languages, respectively. The bracketed numbers indicate the count of languages.}
  \label{tab:multilingual}
\end{table}

\paragraph{Large-Scale Multilingual Translation.}
To further validate whether our method can benefit a broader range of languages, we conducted extensive multilingual translation tasks (99 languages).
Specifically, we utilized OPUS-100 \citep{DBLP:conf/acl/ZhangWTS20} as our training set, which includes bilingual data for 99 languages paired with English, totaling 55M sentences. 
\cref{tab:language_resource} shows the classification of resource levels for these languages.
We trained the models on the X $\to$ En direction and evaluated with Flores-200 dev-test set \citep{DBLP:journals/corr/abs-2207-04672}.  
For our method, we froze the LLM parameters during training and only trained the parameters of the adaptor and decoder. 
Due to the high cost of fine-tuning LLMs on such a large-scale dataset, we did not compare the LLM-SFT model.
As shown in \cref{tab:multilingual}, LaMaTE significantly outperforms the NMT-40-8 model across all resource levels and surpasses the mT5-large model in high and medium language resources. 
This highlights the effectiveness of leveraging LLMs as encoders for multilingual translation, as their rich representations can boost performance.
However, our model underperforms compared to mT5-large in low and very low-resource languages, likely due to limited pre-training data for these languages in the Llama3-base model.
Using a model with stronger multilingual capabilities as the encoder could further improve performance.

\subsection{Efficiency Analysis}

\begin{figure}[b]
    \centering
    \includegraphics[width=\textwidth]{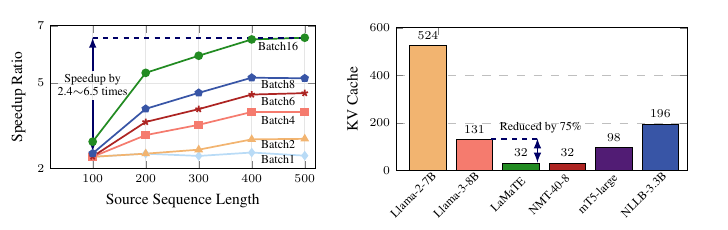}
    \vspace{-0.8cm}
    \caption{Comparison of efficiency: The left chart displays the decoding speedup ratio of LaMaTE versus Llama3-8B under varying source sequence lengths and batch sizes, and the right chart shows the theoretical KV cache size factor for each model.}
    \label{fig:efficiency}
\end{figure}

\cref{fig:efficiency} illustrates the efficiency comparisons between LaMaTE and other models, with their configuration details are provided in \cref{tab:model_configuration}.
The results highlight two key improvements: \textbf{decoding speedup} and \textbf{memory efficiency}.

The left chart in \cref{fig:efficiency} demonstrates that as the source sequence length and batch size increase, LaMaTE achieves a \textbf{2.4× to 6.5× speedup} over Llama3-8B (see \cref{tab:throughput} for details). 
Furthermore, as shown in the right chart, LaMaTE significantly \textbf{reduces KV cache memory by 75\%} compared to Llama3-8B.
This reduction in memory usage enhances scalability, enabling larger batch sizes and longer source sequences. 
For instance, at an 80G memory budget, Llama3-8B encounters out-of-memory issues at a batch size of 24 for source lengths of 300-400, whereas LaMaTE still works normally.
This improvement is particularly beneficial for large-scale inference, where optimizing throughput is crucial for real-world deployment.
Overall, LaMaTE significantly enhances computational efficiency without compromising translation quality, demonstrating its effectiveness in balancing performance and compute resources.

\subsection{Depth vs Performance}

\begin{figure}[hbt]
    \centering
    \includegraphics[width=1.0\linewidth]{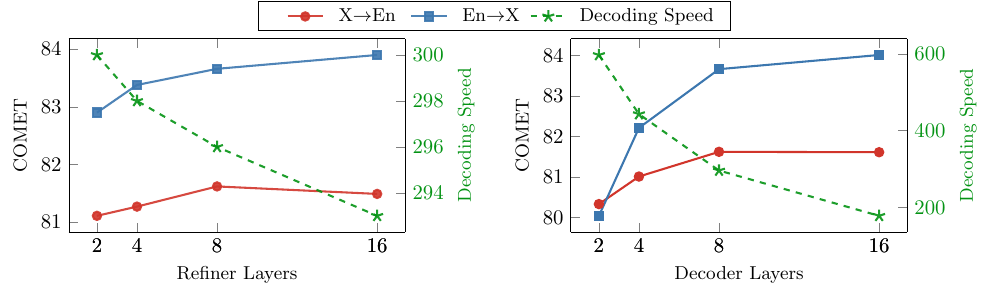}
    \caption{The impact of EncStack and decoder depth on model performance and efficiency.}
    \label{fig.layer_num}
\end{figure}

To analyze how the depth of the EncStack and decoder affects performance and efficiency, we first fix the decoder depth and vary the EncStack depth, then fix the EncStack depth and adjust the decoder depth.
We evaluate translation performance in both En $\to$ X and X $\to$ En directions and measure decoding speed, as shown in \cref{fig.layer_num}.

The left panel presents the effect of increasing the number of EncStack layers. 
The results show that adding more EncStack layers leads to modest improvements in both translation directions, but the gains are relatively small compared to variations in the decoder. 
Additionally, since the EncStack operates only on the encoder side, its impact on decoding efficiency is minimal.

The right panel illustrates the effect of varying decoder depth. 
Unlike the encoder, increasing decoder layers significantly influences both translation performance and efficiency. 
While deeper decoders consistently improve translation quality, particularly in the En $\to$ X direction, they also introduce a notable trade-off: decoding speed drops sharply as depth increases, with a 39\% reduction when moving from 8 to 16 layers. 
This suggests that scaling the encoder offers a more effective trade-off between performance gains and computational efficiency. 
Alternatively, the decoder can be scaled using the Mixture of Experts (MoE) technique, which enables capacity expansion without an increase in inference cost. 
We leave further exploration of these strategies for future work.


\subsection{Ablation Study}
\label{sec:ablation}

\begin{table}[htbp]
  \centering

  \begin{tabular}{lllll}
    \toprule
    \textbf{Aspect} & \textbf{Models} & \textbf{COMET} & \textbf{BLEU} \\
    \midrule
     - & LaMaTE (Ours) &$82.32$ &$33.85$ \\
    \midrule
    \multirow{2}{*}{Training Method} & W/o S2 & $80.07_{(-2.25)}$ & $29.67_{(-4.18)}$ \\
    & W/ S2 \& Frozen LLM &$82.16_{(-0.16)}$ & $33.73_{(-0.12)}$ \\
    \midrule
    \multirow{2}{*}{Adaptor Design} & W/o Layer Fusion &$82.08_{(-0.24)}$ &$33.14_{(-0.71)}$ \\
    & W/o EncStack &$82.02_{(-0.30)}$ &$33.32_{(-0.53)}$ \\
    \midrule
    \multirow{2}{*}{Decoder Variant} & Concat Decoder  &$81.52_{(-0.80)}$ &$31.51_{(-2.34)}$  \\
    & Prefix Decoder  &$81.66_{(-0.66)}$ &$32.38_{(-1.47)}$ \\
    \bottomrule
  \end{tabular}
  \caption{Ablation studies on training methods, adaptor design, and decoder variants. The numbers in the bottom right represent the performance gap relative to the complete model (LaMaTE).}
  \label{tab:ablation_study}
\end{table}
\renewcommand{\arraystretch}{1}

We present ablation studies in \cref{tab:ablation_study}. 
First, training the model with only stage 1 (W/o S2) results in a substantial performance drop. 
Incorporating stage 2 while freezing LLM parameters yields noticeable improvements, but still falls short of LaMaTE's performance.
This underscores the importance of both high-quality data and LLM parameter tuning in maximizing model performance.

For the adaptor, removing the layer fusion leads to a performance drop, indicating that fusing intermediate LLM layers provides a more informative representation than relying solely on the final layer. 
Similarly, removing EncStack results in a slight degradation. 
These findings confirm that the adaptor enhances the native representations of LLMs, rendering them more suitable for the NMT decoder.

Besides, we explore two decoder variants---Concat Decoder and Prefix Decoder---both of which remove the cross-attention layer in the standard decoder.  
For a detailed description of these variants, see Appendix \ref{sec:model_variants}.
As shown in the results, the standard decoder proves superior to alternative designs. 
Together with the observations from \cref{fig:misalignment}, we argue that maintaining cross-attention in the decoder is particularly beneficial for translation tasks that demand strong alignment.

\section{Conclusion}

In this paper, we explore the connection between the two worlds of LLMs and NMT, presenting LaMaTE---a method that leverages LLMs as MT encoders and pairs them with lightweight decoders. 
We design an adaptor better to align LLM's representations for the decoder and propose a two-stage training strategy to develop a universal translation model.
Additionally, we introduce ComMT, a new dataset suite encompassing diverse translation-related tasks, facilitating the development and evaluation of universal translation models. 
Experiments on ComMT demonstrate LaMaTE's impressive performance and generalization ability, while significantly improving computational efficiency---achieving 2.4× to 6.5× faster decoding speeds and reducing KV cache memory usage by 75\%.
We hope this study will provide valuable insights and inspire further exploration into optimizing LLMs and expanding their role in NLP tasks.

\section*{Limitations}
Our approach requires pre-training the decoder's parameters using large-scale bilingual data in the first stage, as the decoder is randomly initialized. 
This process may not be the most efficient. 
Future work could explore initializing these parameters from the encoder or directly leveraging a small pre-trained language model to reduce the state gap between the encoder and decoder.
Additionally, since our decoder is lightweight, it may become a bottleneck when generating translations into many target languages, so expanding its capacity is essential for better performance.
A more effective strategy might involve scaling the decoder's capacity with Mixture-of-Experts (MoE) instead of adding more layers, thereby boosting performance without compromising efficiency.


\bibliographystyle{conf/colm2024_conference}
\bibliography{references}

\begin{appendices}
\onecolumn
\section{Encoder-Decoder vs Decoder-Only}
\label{sec:architecture}

\begin{figure*}[!ht]
    \centering
    \includegraphics[width=0.95\linewidth]{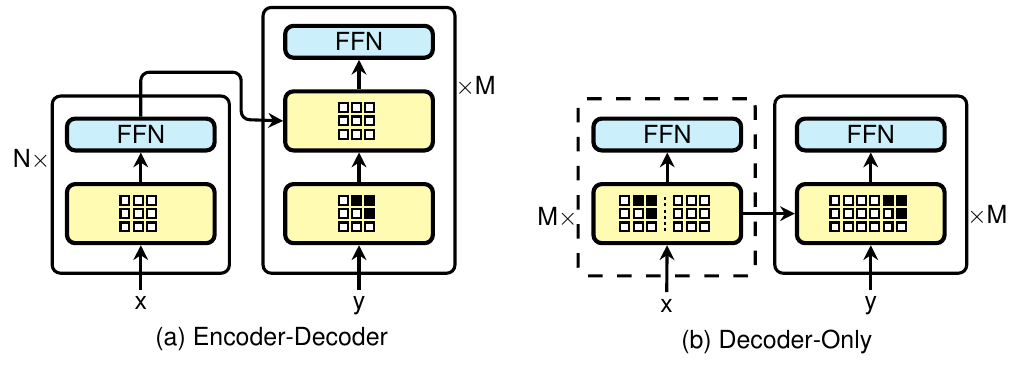}
    \caption{The Encoder-Decoder and Decoder-only architecture.}\label{fig.model-figure}
\end{figure*}

\begin{table*}[htbp]
  \centering
  \begin{adjustbox}{max width=\textwidth}
  \begin{tabular}{lcccccc}
    \toprule
    \multirow{2}{*}{\textbf{Model}} &\multicolumn{2}{c}{\textbf{Connection Type}} &\multicolumn{2}{c}{\textbf{Fuse Type}} &\multirow{2}{*}{\textbf{Src Mask}} &\multirow{2}{*}{\begin{minipage}{4cm} \centering \textbf{Enc-Dec}\\ \centering \textbf{Parameter Sharing}\end{minipage}}   \\
    \cmidrule(lr){2-3} \cmidrule(lr){4-5}
    &\textbf{TopOnly} &\textbf{Layer-Wise} &\textbf{Cross Attention} &\textbf{Concat Attention} & \\
    \midrule
    Encoder-Decoder &\checkmark & &\checkmark & &Fully-visible &\ding{55} \\
    \midrule
    Causal LM & &\checkmark & &\checkmark &Causal &\checkmark \\
    Prefix LM & &\checkmark & &\checkmark &Fully-visible &\checkmark \\
    \bottomrule
  \end{tabular}
  \end{adjustbox}
  \caption{Comparision of different architecture by analyzing their structural elements.}
  \label{tab:model_diffrence}
\end{table*}

Recent advancements in NLP have been profoundly shaped by Transformer models \citep{DBLP:conf/nips/VaswaniSPUJGKP17,radford2018improving,DBLP:conf/naacl/DevlinCLT19}, which have revolutionized both model design and task handling. 
The original Transformer \citep{DBLP:conf/nips/VaswaniSPUJGKP17} was designed for sequence-to-sequence (seq2seq) tasks, utilizing an encoder-decoder architecture where the encoder encodes the input sequence and the decoder generates the output. 
Subsequent models like GPT \citep{radford2018improving} leverage only the Transformer's decoder, omitting the cross-attention layer, for language modeling tasks.
Both architectures are illustrated in \cref{fig.model-figure}.
Although originally designed for different purposes, many NLP tasks can be framed as either seq2seq or language modeling problems \citep{xiao-and-zhu:2025foundations,chang2024efficient}. 

From a macroscopic perspective, the encoder-decoder architecture employs an explicit encoder module to encode the input, after which the decoder generates output based on both the source representation from the encoder's top layer and the target representation from preceding layer. 
Given the source sequence $X$ and target sequence $Y$, this process can be represented as follows \footnote{For simplicity, we omit layer normalization and residual connections in the following formulas and descriptions.}: 

\begin{align}
    X^l& =\operatorname{FFN}(\operatorname{SAtt}(X^{l-1}, M_f)) \label{eq:ed_e} \\
    Y^l& =\operatorname{FFN}(\operatorname{CAtt}(\operatorname{SAtt}(Y^{l-1}, M_c), X^L, M_f)) 
    \label{eq:ed_d}
\end{align}







\noindent where $l$ denotes the layer index, $X^l$ and $Y^l$ represent the representations of the source and target at $l$-th layer, respectively, with $X^0$ and $Y^0$ indicating the embeddings. 
Each layer of the encoder first performs self-attention (SAtt) with a fully-visible mask pattern $M_f$, indicating that all tokens in the source can attend to each other, followed by a feed-forward neural network (FFN). 
The encoder generates its final representation $X^L$ in parallel by stacking L such layers. 
The decoder is similar to the encoder but differs in that: 
(i) due to its autoregressive nature, its self-attention uses a causal mask pattern $M_{c}$, allowing attention only to historical tokens; (ii) it incorporates an additional cross-attention (CAtt) layer to integrate representations from the source, where the mask $M_{f}$ is a fully-visible mask pattern, enabling the target to attend to the entire source.

Contrastingly, the decoder-only architecture processes both the source and target sequences within a single module: 

\begin{equation}
[X^l, Y^l]=\operatorname{FFN}(\operatorname{SAtt}([X^{l-1}, Y^{l-1}], M_c))
\end{equation}

In practice, due to the nature of causal attention, the representation of $X$ is independent of  $Y$.
This allows us to encode $X$ independently to obtain representations at each layer, as shown in \cref{eq:ed_e}.  
Subsequently, the representations of $X^l$ at each layer are concatenated with the corresponding target representations, enabling self-attention computations of the target across both source and target sequences, as shown in \cref{fig.model-figure}(b).
It is important to note that:
(i) when encoding $X$, we can modify its original causal attention mask  
$M_c$ to be fully-visible mask $M_f$, similar to used in the encoder-decoder architecture, thus creating a variant of the CausalLM model known as the prefixLM model \citep{DBLP:conf/nips/00040WWLWGZH19,DBLP:journals/jmlr/RaffelSRLNMZLL20};
(ii) the fusion of source and target information is achieved by computing attention on concatenated representations of both, distinct from cross-attention, which we refer to in this paper as concat attention;
(iii) the interaction between $X$ and $Y$  is Layer-Wise, rather than using only the top-layer representation of X as in the encoder-decoder architecture (TopOnly). 

The overall comparison of these architectures is shown in \cref{tab:model_diffrence}.
In practice, some modern deployment frameworks of LLMs (decoder-only models) explicitly separate encoding (prefilling) and decoding processes across distinct computational resources, making the architecture structurally resemble encoder-decoder models \citep{zhong2024distserve,DBLP:conf/isca/PatelCZSGMB24}.
In this light, the so-called decoder-only model can be considered a variant of the encoder-decoder model, wherein the encoding function is implicitly integrated through shared parameters with the decoder. Conversely, one may view the encoder-decoder model as an extension of PrefixLM, with a more explicit division between encoding and decoding stages.

    \section{ComMT}
\label{sec:app_benchmark}

\subsection{Data Collection, Categorization and Processing}
\label{sec:data_collection}

Our data collection and curation efforts are guided by two key principles: \textbf{ensuring high quality} and \textbf{maintaining diversity} across translation tasks. To this end, we extensively gathered publicly available datasets from the research community, focusing primarily on well-established repositories such as OPUS \footnote{\url{https://opus.nlpl.eu/}}, WMT \footnote{\url{https://www2.statmt.org/wmt23/}}, and Papers with Code \footnote{\url{https://paperswithcode.com/datasets}}. 
These sources have been widely used in MT research.

To uphold data quality, we retained only those datas that had undergone manual annotation, culminating in a collection of over 50 high-quality datasets covering a diverse range of domains, and translation scenarios.
For specialized tasks like document-level translation with limited annotated data, we sampled from resources such as news-commentary.
To further refine our dataset and remove potentially low-quality samples, we employed the COMETkiwi \citep{DBLP:conf/emnlp/ReiSFL20} model for filtering. 
We set the filtering threshold at 0.55 for literature and doc-level translation tasks and 0.75 for all other tasks.
After filtering, we carefully organized and categorized each dataset based on its characteristics according to our classification protocol.
All collected datasets are presented in \cref{tab:benchmark_detail}.
The principles for processing each task are as follows:

\paragraph{General Translation.} 
Datasets lacking distinctive features are classified as general sentence-level tasks. 
We included a small amount of domain-specific data to enhance generalizability, making this task the largest part of ComMT.

\paragraph{Document-level Translation.}
This task requires the model to consider a broader context and capture nuances across sentences \citep{DBLP:journals/csur/MarufSH21}. 
\citet{DBLP:conf/emnlp/JinHMM23} indicates that 3 preceding sentences are usually sufficient to disambiguate most discourse phenomena. 
Based on this finding and practical considerations, we restricted our document-level translation tasks to texts under 500 words and organized the data accordingly.

\paragraph{Domain Translation.}
For the colloquial domain, we focus on informal, non-written text data, mainly from social media platforms, dialogues, and subtitles. For the literature domain, considering the complex discourse phenomena in literary texts, sentence-level translation is unsuitable; thus, we construct this task as multi-sentences form. 
We categorize sentences into varying lengths of [100, 200, 300, 400] words, distributed in proportions of [0.2, 0.3, 0.3, 0.2], to enhance generalizability.

\paragraph{Terminology-constrained Translation.}
Due to the limited availability of terminology translation data, we sampled data and used the B2NERD \citep{DBLP:journals/corr/abs-2406-11192} model to extract term pairs. 
Note that each sample may contain one or more terminology pairs in our data.

\paragraph{Automatic Post-editing.}
We focused on collecting authentic data for this task rather than using synthetic data due to its potential lack of accuracy in representing real post-editing scenarios \citep{DBLP:conf/lrec/NegriTCB18}.

\paragraph{In-context Learning Translation.} 
We designed this task mainly to leverage the inherent In-context learning capabilities of LLMs,  which potentially enable models to adapt on-the-fly.
We extracted 5\% of the data from each task category to create few-shot datasets. 
We structured these into 1-shot, 2-shot, and 3-shot data sets in proportions of [0.3, 0.3, 0.4].

\subsection{Construction of Training and Test Sets}

For train sets, to manage the diverse data sources and prevent excessive data accumulation, we have set a cap of 5,000 samples per dataset. 
For test sets, we ensure that the source language data is "source-original" to closely simulate real-world scenarios and prevent "translationese" effects that could negatively impact evaluation accuracy \citep{DBLP:conf/emnlp/GrahamHK20,DBLP:journals/jair/LaubliCNSST20}.
For general translation tasks, we keep data from WMT22 \citep{DBLP:conf/wmt/AdelaniAABCDFFFGKMMMMSSW22}, WMT23 \citep{DBLP:conf/wmt/KocmiABBDFFFGGH23}, and Flores-200 \citep{DBLP:journals/corr/abs-2207-04672}, and use wmt23 as the default test set.
For most other tasks, we combine samples from multiple sources and strive to maintain a minimum of 500 samples in each test set to ensure reliable evaluation results.
However, some languages currently lack test sets because obtaining the task data is challenging.
In the future, we plan to continue expanding this dataset to support more languages and tasks.

    \section{Detailed Experimental Setups}
\label{sec:setup}
\renewcommand{\arraystretch}{1.1}

\subsection{Datasets and Hyperparameters}
\renewcommand{\arraystretch}{1}
\begin{table*}[htbp]
  \centering
  \begin{adjustbox}{max width=\textwidth}
  \begin{tabular}{lcccc}
    \toprule
    \textbf{Language Pair} &\textbf{De-En} &\textbf{Cs-En} &\textbf{Ru-En} &\textbf{Zh-En} \\  %
    \midrule 
    \textbf{Before Clean} &50.1M &56.3M &39.2M &40.8M \\
    \textbf{After Clean} &46.4M &50.4M &30.8M &33.6M \\
    \bottomrule
  \end{tabular}
  \end{adjustbox}
  \caption{Statistics on the use of parallel data from WMT2023. Note that due to the extensive bilingual data in the En-De CommonCrawl corpus, we only sampled a portion and merged it with other data to create a dataset of 50M. For En-Cs, we excluded the CzEng 2.0 dataset due to licensing issues.}
  \label{tab:parallel_data}
\end{table*}
\renewcommand{\arraystretch}{1.1}

\renewcommand{\arraystretch}{1}
\begin{table*}[h]
\centering
\caption{Hyperparameter configuration during two-stage training and decoding. In the first training stage, we use pure data parallelism because the LLM parameters are frozen. In the second stage, we employed DeepSpeed ZeRO-2 \citep{DBLP:conf/sc/RajbhandariRRH20} for full parameters training.}
\begin{tabular}{p{4cm}p{2cm}p{2cm}}
\hline
\textbf{Hyperparameter} & \textbf{Stage1} & \textbf{Stage2} \\
\hline
Learning Rate & 5e-4 & 2e-5 \\
Adam $\beta$ & (0.9, 0.999) & (0.9, 0.999) \\
LR Scheduler & inverse\_sqrt & cosine \\
Number of Epochs & 1 & 1 \\
Global Batch Size & 2,560 & 384 \\
Train Steps & 30,000 & 1,200 \\
Warmup Ratio & 0.01 & 0.01 \\
Weight Decay & 0.01 & 0.01 \\
\hline
Decoding Method & \multicolumn{2}{c}{beam search} \\
Beam Size & \multicolumn{2}{c}{5} \\
\hline
\end{tabular}
\label{tab:hyperparameters}
\end{table*}
\renewcommand{\arraystretch}{1.1}

\subsection{Beam Search vs Sampling}

\definecolor{mygray}{gray}{.9}
\renewcommand{\arraystretch}{1}
\begin{table*}[htbp]
  \centering
  \begin{adjustbox}{max width=\textwidth}
  \begin{tabular}{llcccccccccc}
    \toprule
    \textbf{Models} &\textbf{Decoding Method} &\textbf{De2En} &\textbf{Ru2En} &\textbf{Zh2En} &\textbf{En2De} &\textbf{En2Cs} &\textbf{En2Ru} &\textbf{En2Zh} &\textbf{Avg.} &\textbf{Speed (tokens/s)}   \\
    \midrule 
    \multirow{2.5}{*}{TowerInstruct-7B} &Beam Search &85.15 &83.18 &80.36 &83.10 &78.88	&85.39 &85.93 &83.16 &114\\
    &Sampling &84.98 &82.49	&79.79 &82.03 &68.97 &84.14 &84.39 &81.15 &167  \\
    \midrule
    \multirow{2.5}{*}{Llama3-8B-SFT} &Beam Search &83.76 &81.66 &76.97 &82.66	&85.80 &82.42
 &80.80 &84.03 &123 \\
    &Sampling &83.48 &80.58 &76.63 &80.01 &81.74 &82.21	&82.03 &80.87 &151\\
    \midrule
    \multirow{2.5}{*}{LaMaTE} &Beam Search &83.90 &81.46 &79.13 &80.73 &86.81 &82.71 &84.41 &82.59 &296 \\
    &Sampling & 82.82 &80.53 &77.97 &80.73 &86.81 &82.71 &84.41 &80.57 &379\\
    \bottomrule
  \end{tabular}
  \end{adjustbox}
  \caption{Comparison of the effectiveness and efficiency of beam search versus sampling.}
  \label{tab:beam}
\end{table*}
\renewcommand{\arraystretch}{1}

While traditional MT research predominantly employs beam search for decoding, LLM-based generation often utilizes sampling strategies for faster and more diverse outputs. 
This raises the question of whether LLMs, when used for translation, should follow the conventional beam search approach or adopt sampling-based decoding. 
To investigate this, we conducted a comparative analysis of beam search (beam size = 5) and sampling (temperature = 0.7, top-k = 50, top-p = 0.8) in terms of translation quality and decoding speed (batch size = 4). 
As shown in \cref{tab:beam}, while beam search is significantly slower, it consistently yields higher translation quality. 
Based on these observations, we adopt beam search as the primary decoding method in this paper.
    \section{Additional Results}
\label{sec:app_result}

\subsection{Detailed Performance on ComMT}
\label{sec:result_app}

We present a detailed performance of each translation task and language pair in the ComMT benchmark in \cref{tab:detail_performance}.

\subsection{Misalignment Evaluation}
To assess the translation misalignment issue, we used three metrics: off-target rate, unaligned source words, and unaligned target words.
Off-target refers to instances where machine-generated translations contain segments from incorrect languages or exhibit code-switching. 
We utilize langdetect \footnote{https://github.com/Mimino666/langdetect} to determine the language of each translation. 
The off-target rate for a translation is calculated by subtracting the probability of the predicted target language from 1. 
We then average this rate across all sentences.
Unaligned source words (USW) refer to words in the source sentence that do not have a corresponding translation in the target sentence. 
Conversely, unaligned target words (UTW) capture instances where words appear in the translation without clear support from the source sentence, indicating potential insertions or hallucinations.
We employ awesome-align \citep{DBLP:conf/eacl/DouN21} to obtain word alignments. 
\cref{tab:misalignment} displays the results of these three indicators.

\begin{table*}[htbp]
  \centering
  \begin{adjustbox}{max width=\textwidth}
  \begin{tabular}{lcrcccccccccc}
    \toprule
    \multirow{2}{*}{\textbf{Models}} &\multicolumn{4}{c}{\textbf{Off-target Rate $\downarrow$}} &\multicolumn{4}{c}{\textbf{USW Rate $\downarrow$}} &\multicolumn{4}{c}{\textbf{USW Rate $\downarrow$}} \\  
     \cmidrule(lr){2-5} \cmidrule(lr){6-9} \cmidrule(lr){10-13} 
    &\textbf{De} &\multicolumn{1}{c}{\textbf{Cs}} &\textbf{Ru} &\textbf{Zh} &\textbf{De} &\textbf{Cs} &\textbf{Ru} &\textbf{Zh} &\textbf{De} &\textbf{Cs} &\textbf{Ru} &\textbf{Zh}  \\
    \midrule
    Llama3-8B-SFT &5.63 &9.45 &2.92 &3.76 &12.03 &17.59	 &17.13 &19.36 &14.16 &14.80 &14.10 &22.59\\
    Tower-7B &6.07 &10.41 &3.49 &2.46 &10.78 &17.65 &15.79 &18.72 &12.39 &15.51 &13.28 &23.93\\
    mT5-large &6.52 &9.14 &2.53 &3.50 &10.00 &15.03 &15.31 &16.88 &11.60 &11.45 &12.51 &18.27\\
    LaMaTE &4.97 &7.93 &2.06 &3.17 &10.74 &16.11 &16.34 &17.57 &11.82 &11.10 &12.61 &17.51\\
    \bottomrule
  \end{tabular}
  \end{adjustbox}
  \caption{Results of the Decoder-only and Encoder-Decoder models on three indicators of misalignment.}
  \label{tab:misalignment}
\end{table*}

\subsection{Comparison of Decoding Speed}

\begin{table*}[htbp]
  \centering
  \begin{adjustbox}{max width=\textwidth}
  \begin{tabular}{lcccccccc}
    \toprule
    \multirow{2.5}{*}{\textbf{\Large Models}} &\multicolumn{3}{c}{\textbf{\Large Decoder-only}} &\multicolumn{4}{c}{\textbf{\Large Encoder-Decoder}}\\
    \cmidrule(lr){2-4} \cmidrule(lr){5-8}
    &\textbf{\Large Llama2-7B} &\textbf{\Large Llama3-8B} &\textbf{\Large Llama2-13B} &\textbf{\Large NMT-40-8} &\textbf{\Large mT5-large} &\textbf{\Large NLLB-3.3B}  &\textbf{\Large LaMaTE (Ours)} \\  %
    \midrule
    \Large Dim &\Large 4096 &\Large 4096 &\Large 5120 &\Large 1024 &\Large 1024 &\Large 2048 &\Large 4096-1024 \\
    \Large Encoder layer &\Large - &\Large - &\Large - &\Large 40 &\Large 24 &\Large 24 &\Large 32-8 \\
    \Large Decoder layer &\Large 32 &\Large 32 &\Large 40 &\Large 8 &\Large 24 &\Large 24 &\Large 8 \\
    \Large Vocab size &\Large 32k &\Large 128k &\Large 32k &\Large 128k &\Large 256k &\Large 256k &\Large 128k \\
    \Large Params &\Large 6.73B &\Large 8.01B &\Large 13.01B &\Large 0.77B  &\Large 1.23B &\Large 3.34B &\Large 8.5B \\
    \cmidrule(lr){1-8}
    \Large \textbf{KV Cache (Kb)} &\Large 524b(s+t) &\Large 131b(s+t) &\Large 819b(s+t) &\Large 32b(s+t) &\Large 98b(s+t) &\Large 196b(s+t) &\Large 32b(s+t) \\
    \bottomrule
  \end{tabular}
  \end{adjustbox}
  \caption{Details on parameter and theoretical KV cache size of the compared models in our work. In the KV Cache, b, s, and t denote the batch size, source sequence length, and target sequence length, respectively. Note that Llama3-8B uses GQA \citep{DBLP:conf/emnlp/AinslieLJZLS23}, resulting in smaller KV cache usage compared to Llama2-7B.}
  \label{tab:model_configuration}
\end{table*}
\renewcommand{\arraystretch}{1}

We first summarize the key architectural details of the evaluated models in \cref{tab:model_configuration}. 
Next, we evaluate the decoding efficiency of these models in \cref{tab:throughput} by comparing their decoding speed across different sequence lengths and batch sizes.
VLLM \citep{DBLP:conf/sosp/KwonLZ0ZY0ZS23} accelerates LLM decoding but has limited beam search support and compatibility issues with some models. 
To ensure fair and consistent evaluation, we use the original Transformers framework across all models.
The results demonstrate that LaMaTE consistently outperforms Llama3-8B in decoding speed, particularly as batch sizes and input lengths increase, achieving a speedup ranging from 2.4× to 6.5×.

\definecolor{mygray}{gray}{.9}
\renewcommand{\arraystretch}{1}
\begin{table*}[htbp]
  \centering
  \begin{adjustbox}{max width=\textwidth}
  \begin{tabular}{@{\extracolsep{\fill}}clccccccc>{\columncolor{mygray}}c}
    \toprule
    \multirow{2.5}{*}{\textbf{Batch Size}} & \multirow{2.5}{*}{\textbf{Length}} &\multicolumn{3}{c}{\textbf{Decoder-only}} & \multicolumn{4}{c}{\textbf{Encoder-Decoder}} & \multirow{2}{*}{\textbf{Speedup$\uparrow$}} \\
    \cmidrule(lr){3-5} \cmidrule(lr){6-9}
    ~ & ~ &\textbf{Llama2-7B} &\textbf{Llama3-8B} &\textbf{Llama2-13B} &\textbf{NMT-40-8} &\textbf{mT5-large} &\textbf{NLLB-3.3B}  &\textbf{LaMaTE (Ours)} & ~ \\  
    \midrule
    \multirow{5}{*}{1} & 0-100 & 35 & 32 & 28 & 170 & 43 & 48 & 77 & 2.41 \\
~ & 100-200 & 33 & 33 & 23 & 167 & 44 & 42 & 83 & 2.52 \\
~ & 200-300 & 31 & 34 & 21 & 162 & 43 & 38 & 83 & 2.44 \\
~ & 300-400 & 29 & 32 & 19 & 160 & 42 & 36 & 82 & 2.56 \\
~ & 400-500 & 29 & 33 & 18 & 160 & 43 & 36 & 81 & 2.45 \\
    \midrule
    \multirow{5}{*}{2} & 0-100 & 67 & 63 & 51 & 310 & 84 & 77 & 152 & 2.41 \\
~ & 100-200 & 53 & 62 & 34 & 294 & 83 & 63 & 156 & 2.52 \\
~ & 200-300 & 45 & 58 & 29 & 289 & 81 & 55 & 154 & 2.66 \\
~ & 300-400 & 38 & 52 & 24 & 281 & 80 & 52 & 157 & 3.02 \\
~ & 400-500 & 38 & 51 & 24 & 283 & 79 & 51 & 155 & 3.04 \\
    \midrule
    \multirow{5}{*}{4} & 0-100 & 114 & 123 & 76 & 515 & 164 & 181 & 296 & 2.41 \\
~ & 100-200 & 57 & 87 & 35 & 472 & 141 & 112 & 276 & 3.17 \\
~ & 200-300 & 47 & 77 & OOM & 453 & 133 & 94 & 272 & 3.53 \\
~ & 300-400 & 38 & 66 & OOM & 421 & 123 & 82 & 263 & 3.98 \\
~ & 400-500 & 38 & 65 & OOM & 422 & 123 & 79 & 259 & 3.98 \\
    \midrule
    \multirow{5}{*}{6} & 0-100 & 144 & 172 & 90 & 668 & 230 & 217 & 416 & 2.42 \\
~ & 100-200 & 59 & 100 & OOM & 565 & 183 & 116 & 363 & 3.63 \\
~ & 200-300 & 49 & 87 & OOM & 549 & 171 & 95 & 355 & 4.08 \\
~ & 300-400 & 39 & 73 & OOM & 495 & 155 & 84 & 335 & 4.59 \\
~ & 400-500 & 39 & 72 & OOM & 500 & 154 & 83 & 334 & 4.64 \\
    \midrule
    \multirow{5}{*}{8} & 0-100 & 162 & 210 & 100 & 789 & 289 & 232 & 532 & 2.53 \\
~ & 100-200 & 59 & 109 & OOM & 643 & 216 & 119 & 446 & 4.09 \\
~ & 200-300 & OOM & 93 & OOM & 617 & 198 & 99 & 432 & 4.65 \\
~ & 300-400 & OOM & 78 & OOM & 553 & 177 & 87 & 404 & 5.18 \\
~ & 400-500 & OOM & 77 & OOM & 556 & 175 & 82 & 397 & 5.16 \\
    \midrule
    \multirow{5}{*}{16} & 0-100 & 167 & 278 & 95 & 956 & 446 & 281 & 816 & 2.94 \\
~ & 100-200 & OOM & 118 & OOM & 728 & 269 & OOM & 631 & 5.35 \\
~ & 200-300 & OOM & 99 & OOM & 690 & 244 & OOM & 589 & 5.95 \\
~ & 300-400 & OOM & 82 & OOM & 609 & 207 & OOM & 535 & 6.52 \\
~ & 400-500 & OOM & 81 & OOM & 602 & 206 & OOM & 533 & 6.58 \\
    \midrule
    \multirow{5}{*}{24} & 0-100 & 130 & 308 & OOM & 976 & 550 & 307 & 1050 & 3.41 \\
~ & 100-200 & OOM & 121 & OOM & 739 & 263 & OOM & 728 & 6.02 \\
~ & 200-300 & OOM & 102 & OOM & 708 & 250 & OOM & 667 & 6.54 \\
~ & 300-400 & OOM & OOM & OOM & 624 & OOM & OOM & 594 & - \\
~ & 400-500 & OOM & OOM & OOM & 619 & OOM & OOM & 593 & - \\
    \bottomrule
  \end{tabular}
  \end{adjustbox}
  \caption{Evaluate model decoding speed (tokens/s) across various batch sizes and source sequence lengths. Speedup indicates the decoding speedup ratio of LaMaTE versus Llama3-8B.}
  \label{tab:throughput}
\end{table*}
\renewcommand{\arraystretch}{1}

\subsection{Comparison of Decoder Variants}
\label{sec:model_variants}

The decoder of the original transformer utilizes cross attention to integrate the encoder's representation, as shown in \cref{eq:ed_d}. 
We refer to this standard decoder Cross Decoder. 
We propose two variants of the decoder that omit the cross-attention layer. 

The first variant, referred to as the Concat Decoder, handles the encoder’s representation $H_E$ by incorporating it directly into the self-attention layers of the decoder.
Specifically, in the self-attention computation, the keys and values are computed from $[H_E, Y^{l-1}]$, while the queries are derived solely from $Y^{l-1}$. 
Thus, this allows target tokens to integrate source and target information within a single attention computation:

\begin{equation}
Y^l=\operatorname{FFN}(\operatorname{SAtt}([H_E, Y^{l-1}]))
\end{equation}

The second variant, taking inspiration from recent research in multimodal language model \citep{DBLP:conf/nips/LiuLWL23a}, referred to as Prefix Decoder, where the encoder representations are concatenated directly with the decoder’s embeddings $Y_0$ before being fed to the upper decoder layers:
\begin{equation}
Y^l=\operatorname{FFN}(\operatorname{SAtt}(Y^{l-1})), \quad Y^0=[H_E, Y^0]
\end{equation}
To preserve the bidirectional nature of source representations within the decoder, we adopt a masking strategy similar to PrefixLM, ensuring that source-side tokens retain their bidirectional feature.

\begin{figure*}[!ht]
    \centering
    \includegraphics[width=1.0\linewidth]{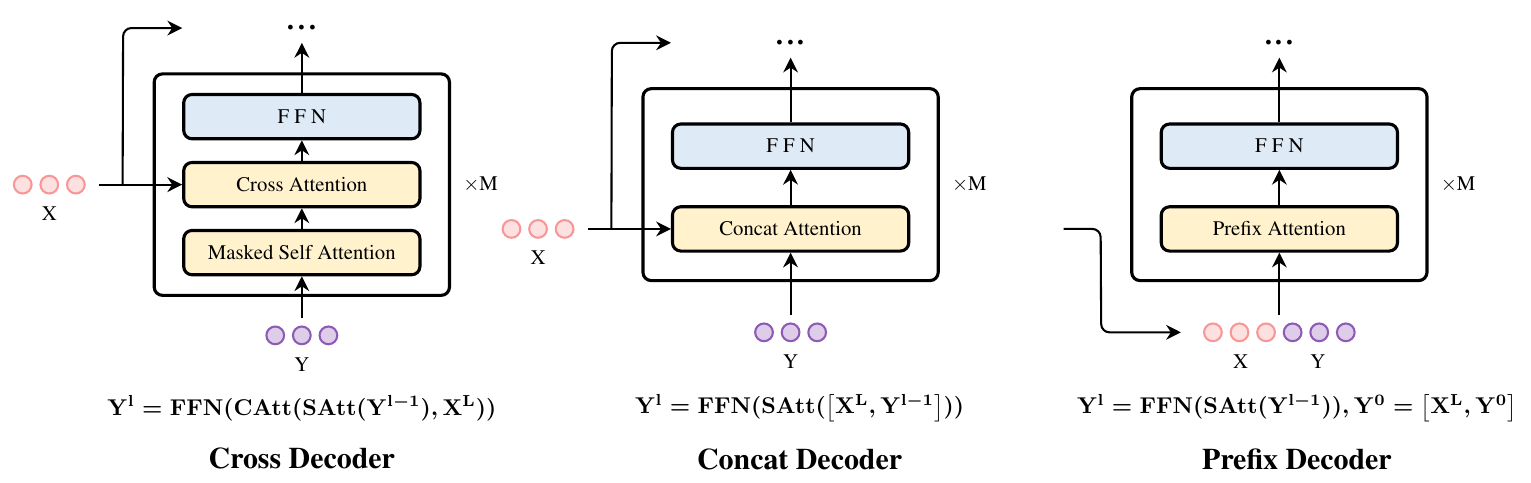}
    \caption{Three variants of decoders: Cross Decoder is the standard decoder, while Concat Decoder and Prefix Decoder remove the cross-attention sublayer, integrating source information through self-attention and early fusion methods, respectively.}
    \label{fig.decoder_variants}
\end{figure*}

A comparative overview of the three variants is presented in \cref{fig.decoder_variants}.
\cref{tab:model_variant_detail} displays their performances.
As shown, the Cross Decoder, i.e., the standard decoder, achieves the best overall performance.

\renewcommand{\arraystretch}{3}
\begin{table*}[htbp]
  \centering
  \begin{adjustbox}{max width=\textwidth}
\begin{tabular}{lp{4.5cm}cccccccccc}
    \toprule
    \multirow{2}{2.5cm}{\textbf{\myHuge Decoder\textcolor{white!50}{y}}} &\multirow{2}{4.5cm}{\textbf{\myHuge Training \textcolor{white!50}{1} Stage}} &\multicolumn{2}{c}{\textbf{\myHuge De}} &\multicolumn{2}{c}{\textbf{\myHuge Cs}} &\multicolumn{2}{c}{\textbf{\myHuge Ru}} &\multicolumn{2}{c}{\textbf{\myHuge Zh}} &\multicolumn{2}{c}{\textbf{\myHuge Avg.}}\\  
     \cmidrule(lr){3-4} \cmidrule(lr){5-6} \cmidrule(lr){7-8} \cmidrule(lr){9-10} \cmidrule(lr){11-12}
    & &\textbf{\myHuge COMET} &\textbf{\myHuge BLEU} &\textbf{\myHuge COMET} &\textbf{\myHuge BLEU} &\textbf{\myHuge COMET} &\textbf{\myHuge BLEU} &\textbf{\myHuge COMET}  &\textbf{\myHuge BLEU} &\textbf{\myHuge COMET} &\textbf{\myHuge BLEU} \\
    \midrule
   \multirow{2}{*}{\myHuge Cross Decoder} &\myHuge \textcolor{white!50}{say} 1 &\myHuge 79.93 	&\myHuge 31.68 	&\myHuge 85.31 	&\myHuge 33.22 	&\myHuge 79.38 	&\myHuge 26.67 	&\myHuge 78.64 	&\myHuge 29.92 	&\myHuge 80.07 	&\myHuge 29.67   \\
    &\myHuge \textcolor{white!50}{say} 2 &\myHuge 82.22 	&\myHuge 39.50 	&\myHuge 86.54 	&\myHuge 35.34 	&\myHuge 81.71 	&\myHuge 28.61 	&\myHuge 81.46 	&\myHuge 34.00 	&\myHuge 82.32 	&\myHuge 33.85   \\\cmidrule(lr){1-12}
    \multirow{2}{*}{\myHuge Concat Decoder} &\myHuge \textcolor{white!50}{say} 1 &\myHuge 79.45 	&\myHuge 30.50 	&\myHuge 85.02 	&\myHuge 32.89 	&\myHuge 79.45 	&\myHuge 26.40 	&\myHuge 78.99 	&\myHuge 30.15 	&\myHuge 80.01 	&\myHuge 29.33 \\
   &\myHuge \textcolor{white!50}{say} 2 &\myHuge 79.89 	&\myHuge 32.38 	&\myHuge 85.97 	&\myHuge 34.04 	&\myHuge 81.63 	&\myHuge 28.47 	&\myHuge 81.14 	&\myHuge 33.49 	&\myHuge 81.52 	&\myHuge 31.51  \\\cmidrule(lr){1-12}
    \multirow{2}{*}{\myHuge Prefix Decoder} &\myHuge \textcolor{white!50}{say} 1 &\myHuge 79.12 	&\myHuge 30.04 	&\myHuge 85.37 	&\myHuge 32.73 	&\myHuge 79.71 	&\myHuge 26.73 	&\myHuge 79.24 	&\myHuge 30.21 	&\myHuge 80.11 	&\myHuge 29.25  \\
\bottomrule
  \end{tabular}
  \end{adjustbox}
  \caption{A comparison of decoder variants.}
  \label{tab:model_variant_detail}
\end{table*}
\renewcommand{\arraystretch}{1}

\begin{table*}[htbp]
  \centering
  \begin{adjustbox}{max width=\textwidth}
  \begin{tabular}{ll}
    \toprule
    \textbf{Resource} & \textbf{Languages} \\  %
    \midrule 
        & Arabic (ar), German (de), Spanish (es), French (fr), Italian (it), \\
    High & Japanese (ja), Dutch (nl), Polish (pl), Portuguese (pt), Russian (ru), \\
    (>1\%)  & Turkish (tr), Chinese (zh) \\
    \specialrule{0em}{0pt}{1pt}
    \specialrule{0.05em}{1pt}{1pt}
    \specialrule{0em}{0pt}{1pt}
        & Bulgarian (bg), Bengali (bn), Catalan (ca), Czech (cs), Danish (da), \\
        & Modern Greek (el), Estonian (et), Persian (fa), Finnish (fi), Hindi (hi), \\
    Medium & Hungarian (hu), Indonesian (id), Korean (ko), Lithuanian (lt), Latvian (lv), \\
    (>0.1\%)  & Norwegian (no), Romanian (ro), Slovak (sk), Slovene (sl), Swedish (sv), \\
\bottomrule
  \end{tabular}
  \end{adjustbox}
  \caption{We categorized 99 languages in OPUS-100 \citep{DBLP:conf/acl/ZhangWTS20} into four resource levels—--high, medium, low, and very low---based on the proportion of data available on the internet. }
  \label{tab:language_resource}
\end{table*}

\definecolor{mygray}{gray}{.9}
{
\onecolumn
\renewcommand{\arraystretch}{1.1}
\newcommand{\yourLarge}{\fontsize{9}{9}\selectfont}
\begin{center}
\begin{scriptsize}
\setlength{\tabcolsep}{1pt}
\end{CJK}
}

\end{appendices}

\end{document}